\documentclass[lettersize,journal]{IEEEtran}
\usepackage{amsmath,amsfonts, amssymb}
\usepackage{algorithmic}
\usepackage{algorithm}
\usepackage{amsthm} 
\usepackage{makecell}
\usepackage{array}
\usepackage[caption=false,font=normalsize,labelfont=sf,textfont=sf]{subfig}
\usepackage{textcomp}
\usepackage{stfloats}
\usepackage{url}
\usepackage{pifont}
\usepackage{verbatim}
\usepackage{graphicx}
\usepackage{hyperref}
\usepackage{units}
\usepackage{multirow}
\usepackage{cite}
\usepackage{balance}
\begin{document}

\title{MvT: Event-based Multi-view Projection for Multiple Object Tracking}

\author{Muxi Zha, Banglei Guan, Minzu Liang, Zibin Liu, Yang Shang, Qifeng Yu, and Laurent Kneip
    \thanks{This work was supported by the National Natural Science Foundation of China under Grant 12372189 and the Science and Technology Innovation Program of Hunan Province 2025RC1045.}
    \thanks {Muxi Zha, Banglei Guan, Zibin Liu, Yang Shang, and Qifeng Yu are with the College of Aerospace Science and Engineering, National University of Defense Technology, Changsha, 410073, Hunan, China, and with the Hunan Provincial Key Laboratory of Image Measurement and Vision Navigation, Changsha, 410073, Hunan, China. (Corresponding authors: Banglei Guan; Yang Shang).}
    \thanks{Minzu Liang is with the College of Science, National University of Defense Technology, Changsha 410073, China.}
    \thanks{Laurent Kneip is with the Mobile Perception Laboratory, School of Information Science and Technology, ShanghaiTech University, Shanghai 201210, China.}
    \thanks{Digital Object Identifier 10.1109/TIP.2026.3726353}
}

\IEEEpubid{\begin{minipage}{\textwidth}\ \centering 
    1941-0042 \copyright~2026 IEEE. All rights reserved, including rights for text and data mining, and training of artificial intelligence and \\
similar technologies. Personal use is permitted, but republication/redistribution requires IEEE permission.\end{minipage}}

\maketitle

\begin{abstract}
    Event cameras are increasingly used for Multiple Object Tracking (MOT), but their asynchronous event output often requires specialized methods. Existing processing methods primarily follow two paradigms, pseudo-frames and event-by-event. The former is the prevailing approach since its data format aligns with images, making image-based techniques applicable. However, it suffers from tracking failures when trajectories overlap or are spatially close on pseudo-frames. Facing this challenge, we propose a multi-view pipeline, Multi-view Tracking (MvT), which preserves the 2D data format to leverage image-based techniques directly while introducing additional spatio-temporal views to resolve tracking ambiguities in a single view. MvT comprises a Multi-view Projection (MvP) module and a Multi-view Fusion (MvF) stage. MvP encodes events into three complementary spatio-temporal views while mitigating the pattern discretization. Within MvF, multi-view results are unified into a 3D coordinate system, and tracklets are associated through an optimization model subject to specific criteria combination. Evaluations on four datasets, including our self-collected Small Objects Dataset (SOD), show that MvT seamlessly integrates image-based methods and outperforms existing non-learning and learning trackers in generalized scenarios, and effectively resolves the single-view tracking ambiguities. Being training-free, MvT is applicable when ground-truth annotation is infeasible, thereby highlighting its practical, data-efficient potential. Code is available at \url{https://github.com/zhazhabiu/MvTracking}.
\end{abstract}


\section{Introduction}
\label{section:Introd}

\begin{figure}[t!]
	\begin{center}
		\centerline{\includegraphics[width=.9\linewidth]{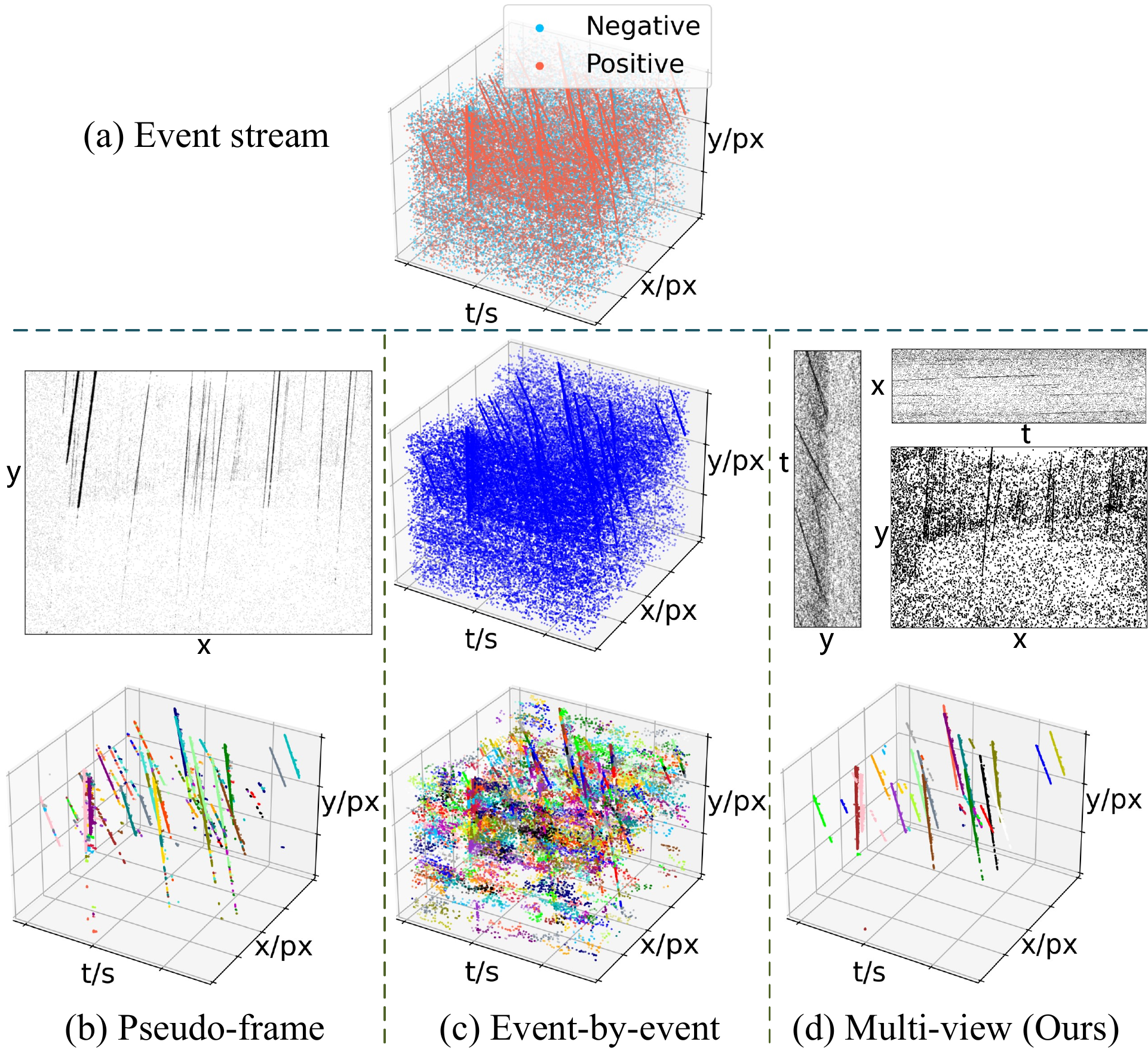}}
		\vspace{-0.3cm}
		\caption{Comparison of different event processing approaches. (a) Asynchronous event streams of the Raindrops sequence in SOD. (b)-(d) The first row shows data representations (0.3$s$ period) of pseudo-frames (event images~\cite{gallego2017accurate} here), event-by-event processing, and our multi-view representation. The second row shows the corresponding tracking results from: our single-view framework variant with region growing, E-MS~\cite{barranco2018real}, and the original MvT+RG. Different colors indicate distinct trajectories.}
		\label{fig1}
		\vspace{-0.8cm}
	\end{center}
\end{figure}

\IEEEPARstart{M}{ultiple} Object Tracking (MOT) serves as a fundamental tool for various applications, such as pose estimation~\cite{Guan2025IJCV, TPAMI_Guan} and autonomous driving assistance~\cite{hu2019joint}. Image-based trackers excel in precision but struggle under challenging conditions (e.g., ground-based UAV observation). Event cameras offer a promising alternative~\cite{brandli2014240} with high dynamic range, low latency, and fast response to stimuli in these extreme environments. However, their asynchronous event output is fundamentally different from images, as shown in Fig.~\ref{fig1}(a). Specialized event-based tracking methods have been developed~\cite{chen2019asynchronous, chen2020end, barranco2018real}, but they mainly leverage the high dynamic range and low latency of event cameras to achieve real-time tracking. With the special motion sensitivity of event cameras, this paper focuses on small object tracking, which is more challenging than standard MOT due to issues such as small sizes and similar appearance features.

\IEEEpubidadjcol
Each output event contains four elements, spatial coordinates $(x, y)$, timestamps $t$, and polarity $p$. Current event-based tracking methods often process them in two formats, pseudo-frames~\cite{gallego2017accurate, gallego2018unifying} and event-by-event~\cite{barranco2018real, ni2012asynchronous1}, as depicted in Fig.~\ref{fig1}(b) and (c). The first approach generates pseudo-frames from events, including event images, voxel-grids, time surfaces, etc. Since this format aligns with images, which are intuitive and convenient to process, the majority of event-based tracking works~\cite{chen2019asynchronous, chen2020end, zhuCrossModalOrthogonalHighRank2023, zhangSpikingNeuralNetworks2025} are built upon it. Compared to real images, most pseudo-frame representations lack dense color and texture information, which serves as critical appearance cues for tracking. Although some studies~\cite{paredes-vallesBackEventBasics2021, scheerlinckFastImageReconstruction2020} have successfully utilized event polarity, which encodes the luminance gradient fields, to recover texture and intensity images, commonly used event images lack such rich cues. Furthermore, they simply collapse the temporal dimension of events, which can lead to tracking failures when object trajectories overlap or intersect in the image plane.

Event-by-event processing preserves the original sequential temporal nature of events. Beyond emerging Spiking Neural Networks (SNN)~\cite{zhangSpikingNeuralNetworks2025, zhangSpikingTransformersEventbased2022}, clustering techniques~\cite{schraml2010spatio, bolten2019application, barranco2018real, hu2022ecdt} are used widely since they are training-free. These methods regard events as point clouds distributed in t-x-y space. However, unlike standard 3D point clouds with homogeneous axis units in z-x-y, the heterogeneous units of t-x-y axes complicate parameter selections. Specifically, first, the high temporal resolution and asynchronous nature cause a relatively sparser distribution along t-axis compared to the spatial dimensions, thus requiring disparate clustering parameters for different axes. Second, the temporal distribution of an event trajectory is velocity-dependent, rendering it hard to determine parameters for objects with diverse motion patterns. Though this format avoids the tracking ambiguities in above pseudo-frames, these issues limit its generalizability.

Given the processing ease and straightforward parameter determination of pseudo-frames, alongside the rich spatio-temporal information of event-by-event processing for trajectory disambiguation, an open problem arises. \textit{How can the tracking ambiguities inherent in single pseudo-frames be resolved?} Considering the difficulty in obtaining ground truth annotations from events, we move away from prevailing supervised network designs.

\begin{figure*}[!t]
	\begin{center}
		\vspace{-0.3cm}
		\centerline{\includegraphics[width=.9\linewidth]{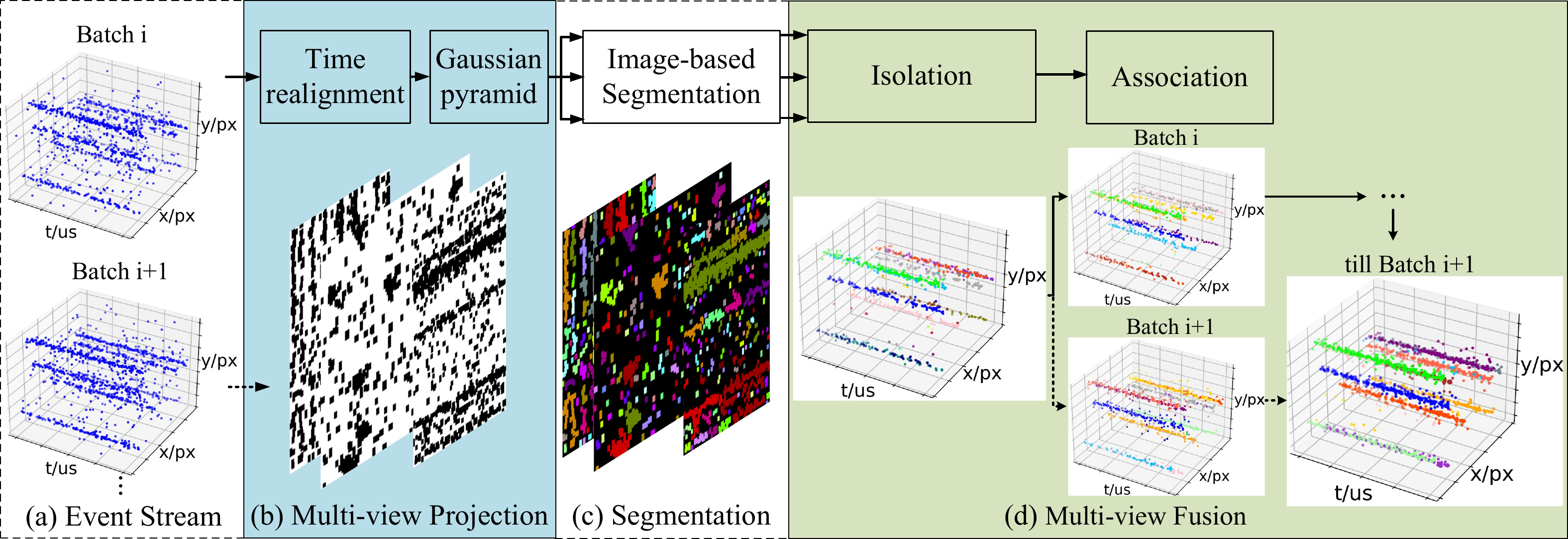}}
		\vspace{-0.3cm}
		\caption{Overview of our MvT pipeline and visualization of main stages, taking an image-based segmentation method as an example. (a) An event stream is divided into blocks using a sliding window and then fed into the processing stages one by one. (b) and (d) detail the key components of MvT, MvP and MvF, respectively. MvP generates multiple views through time-realigned events. In MvF, complete trajectories are recovered from the segmentation results of multiple views. (c) denotes the image-based segmentation stage.}
		\label{fig3}
		\vspace{-0.6cm}
	\end{center}
\end{figure*}

In this paper, we provide a novel generalizable framework MvT, to integrate existing image-based methods into event-based tracking task. To ensure compatibility with image-based methods, MvT preserves the pseudo-frame format. The overall pipeline is shown in Fig.~\ref{fig3}. It encodes events into a multi-view representation through its component, MvP. The generated views include a spatial view, XOY, and two spatio-temporal views, YOT and TOX. This representation overcomes the information limitation of a single pseudo-frame. Image-based detection/segmentation methods are then applied to views. The other component of MvT, named MvF, fuses results from each view and maps them back into the original t-x-y coordinates. If there are detection or segmentation errors in some views, MvF can leverage complementary information provided by other views to correct results through an isolation scheme. Subsequently, an association stage is performed on the corrected results. To accommodate the modality of event data, we design three association criteria based on the motion consistency of event trajectories.

In experiments, we integrate both non-learning and learning-based methods into MvT, and compare them against carefully devised pseudo-frame and event-by-event methods. Evaluations are conducted on four datasets, including Event Camera Dataset (ECD)~\cite{mueggler2017event}, Bee Swarm Dataset (BEES)~\cite{apps2025asynchronous}, EV-UAV~\cite{chenEventbasedTinyObject2025}, and our self-collected Small Objects Dataset (SOD). Except for ECD, the remaining three datasets cover small-sized objects and various motion patterns.
The main contributions of this paper are as follows:
\begin{itemize}
	\item We propose MvT, a generalized and training-free event-based MOT framework that enables the direct application of image-based methods while preserving the temporal information of events for tracking.
	\item We devise an association model comprising three criteria, corresponding to three cases derived from short-term and long-term event trajectory relationships. These criteria can be flexibly combined to handle diverse association scenarios involving both normal-sized and small-sized objects.
	\item Extensive experiments validate the effectiveness and compatibility of MvT with various image-based detection and segmentation methods, as well as the efficacy of the multi-view representation.
\end{itemize}

\section{Related Work}
\label{section:RW}
We provide a review of event-based tracking methods, covering both pseudo-frames and event-by-event paradigms. We then introduce the multi-view representation in image processing and spike-based coding, which collectively inspire the design of our event representation.

\subsection{Pseudo-frame Tracking}

Pseudo-frames refer to 2D images constructed from events, such as event images~\cite{gallego2017accurate}, Image of Warped Events (IWE)~\cite{gallego2018unifying}, and voxel-grids~\cite{zhang2021object, zhangSpikingTransformersEventbased2022}, with IWE being a well-known form that accumulates events warped by motion models. This type of frame is typically processed using maximization optimization, such as Entropy Maximization (EM)~\cite{nunes2020entropy, nunes2021robust}, Expectation Maximization-Iterative Closest Point (EM-ICP)~\cite{zhu2017event}, Contrast Maximization (CM)~\cite{gallego2018unifying, gallego2017accurate, stoffregen2019event, shiba2022secrets, stoffregen2018simultaneous}, and Spatio-temporal Consistency Maximization~\cite{liu2021spatio-temporal}. For example, Zhu et al.~\cite{zhu2017event} introduce an optical flow method based on Expectation Maximization that maximizes all correlation expectations of soft data associations for feature tracking. In other pseudo-frame-based works, Zhang et al.~\cite{zhang2021object} design cross-domain attention schemes that combine the information from RGB frames and voxel-grid-based event frames to guide object tracking. 
Zhang et al.~\cite{zhangSpikingTransformersEventbased2022} then generate positive-negative event frames and feed them into a transformer module for global spatial information, as well as a spiking neural network (SNN) for temporal clues. With these clues, a single object is accurately detected and tracked. Since event images cannot fully utilize temporal information, researchers have explored diverse pseudo-frames that integrate more temporal information. For example, the Surface of Active Events (SAE)~\cite{benosman2013event} stores timestamps as pixel values and has been used to estimate optical flow. Subsequently, a Time-Surface map, which stores exponentially decayed timestamps, was introduced as a prototype of event-based object recognition tasks~\cite{lagorce2016hots} and has since been extended as a popular representation in motion estimation~\cite{huang2023progressive, sironi2018hats}. Chen et al.~\cite{chen2019asynchronous, chen2020end} propose an Adaptive Time-Surface with Linear Time Decay (ATSLTD) frame representation and a Time-Surface with Linear Time Decay (TSLTD) representation. Almareafi et al.~\cite{almatrafi2020distance} introduce a prototype target texture that captures temporal distance variations in events. These representations retain temporal information by encoding timestamps into pixel values, but they may lose other details, like the number of triggers at each pixel. Mitrokhin et al.~\cite{mitrokhin2018event} propose using a Time-image to assist event images in jointly estimating motions.

\subsection{Event-by-event Tracking}
To fully leverage the low latency of event cameras, many asynchronous processing methods bypass frame generation, categorizing object tracking into one-stage and two-stage approaches. In one-stage methods, deep learning models, like Vision Transformers (ViT)~\cite{zhu2023cross} and SNNs serve as a supervised solution mainly for Single Object Tracking (SOT)~\cite{zhangSpikingNeuralNetworks2025, zhangSpikingTransformersEventbased2022}, while unsupervised clustering offers a more general alternative applicable to MOT without extensive training~\cite{schraml2010spatio, bolten2019application, ni2012asynchronous1}. However, varying target speeds cause unstable performance due to non-fixed clustering criteria. Two-stage pipelines separate detection and tracking. The detection stage utilizes not only clustering methods~\cite{rodriguez2020asynchronous, barranco2018real, hu2022ecdt} but also some visual feature-based methods~\cite{2024_Wang_AEB_Tracker_TRO, ni2012asynchronous2, alzugaray2020haste}. The tracking stage associates tracklets across batches similar to image-based tracking. Several methods have been developed, including searching features or templates in the neighborhood~\cite{rodriguez2020asynchronous, alzugaray2020haste}, matching target descriptors across adjacent time steps~\cite{hu2022ecdt}, and employing Kalman filters~\cite{barranco2018real, 2024_Wang_AEB_Tracker_TRO, borer20144d}.

\subsection{Multi-view based works}
Event data can be regarded as point clouds in the T-X-Y space. In processing point clouds, leveraging multiple views is a common strategy to obtain more information than a single view provides. ``Multiple views'' typically carries two meanings, views at different timestamps (video sequences) and views from different perspectives. In our work, we focus on the latter, which is more relevant to event data. Among multi-view forms, orthographic views are particularly effective, being used for reconstruction via geometric constraints~\cite{kuo1998reconstruction}, and for classification using morphological information~\cite{guo2020deep}. This principle persists in modern neural representations. For example, the Triplane representation in Neural Radiance Fields (NeRF)~\cite{chan2022efficient, khatib2024trinerflet} uses three 2D feature planes, facilitating the application of 2D convolutional networks. Multi-view representation is also used in spike-based coding. Thorpe et al.~\cite{thorpeSpikebasedStrategiesRapid2001} propose Rank Order Coding, which involves projecting spatio-temporal data onto three 2D planes (X-T, Y-T, and X-Y). This projection scheme has been explored in neuroscience~\cite{adelson1983spatio-temporal} and applied to image processing~\cite{bollesEpipolarplaneImageAnalysis1987}, establishing a foundational method for converting temporal event streams into informative 2D maps.

\section{Method}
\label{section:Method}

For a better understanding of this work, we start with a motivation analysis in Sec.~\ref{section:motivation} to guide the design of our framework. A brief introduction to MvT is provided in Sec.~\ref{section:MvT}, followed by detailed information about MvP (Sec.~\ref{section:MvP}) and MvF (Sec.~\ref{section:MvF}).

\begin{figure}[!t]
    \begin{center}
        \centerline{\includegraphics[width=.9\linewidth]{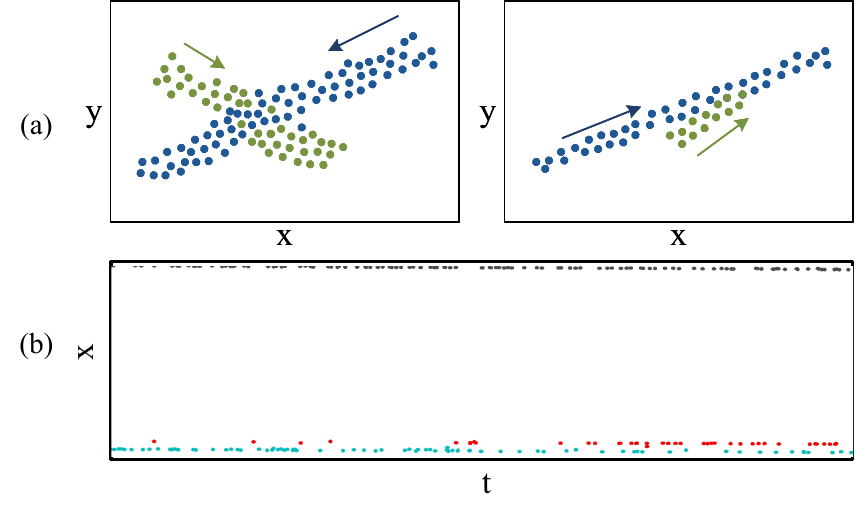}}
        \vspace{-0.25cm}
        \caption{Illustrations of challenges in current event-based tracking paradigms. (a) In pseudo-frame representations, overlapping trajectories (green and blue) may lead to erroneous tracking. Arrows indicate the direction of motion. (b) In the TOX plane, each trajectory consists of a series of discrete events. Objects moving at different velocities (red, blue, and dark gray) generate trajectories with distinct temporal distributions, which necessitate different processing parameters.}
        \label{fig2}
        \vspace{-0.6cm}
    \end{center}
\end{figure}

\subsection{Motivation}
\label{section:motivation}
While image-based detection or segmentation methods have demonstrated remarkable success, their direct application to event-based tracking remains challenging due to modality shift. As analyzed in Sec.~\ref{section:Introd}, existing event-based tracking approaches face two primary challenges.
(1) Single pseudo-frame representations contain limited spatio-temporal information. When objects overlap or their trajectories intersect in the 2D projection, these representations fail to distinguish them (Fig.~\ref{fig2}(a)).
(2) The spatio-temporal units (t-x-y) are not homogeneous. The inherent mismatch between temporal ($\mu s$) and spatial (pixel) units, combined with the asynchronous triggering mechanism, results in sparse and non-uniform event distributions. As illustrated in Fig.~\ref{fig2}(b), varying target speeds lead to distinctly different event densities (total event counts per unit of recording period) along trajectories, thereby posing difficulties for parameter selection. We summarize this as \textbf{Claim~\ref{C1}}. The proof is provided in the supplementary material.

Moreover, although the high temporal resolution of event cameras enables tracking to be treated as a trajectory segmentation problem, it introduces a side effect, discretization. From a macroscopic perspective, trajectories can be regarded as continuous 3D lines or volumes. However, at the micro-time scale, they comprise discretely distributed events (Fig.~\ref{fig2}(b)). In this work, we consider both temporal discretization resulting from sparse event triggering and spatial discretization caused by the physical artifact of pixel-based sensor arrays.

For the first challenge, our multi-view representation adopts the 2D view format that enables the application of image-based algorithms while using multiple views to preserve essential spatio-temporal information for trajectory discrimination.
As for the second challenge and the discretization issue, we introduce a time realignment method designed for asynchronously triggered events, alongside a Gaussian pyramid, to alleviate the tracklet pattern discontinuities caused by spatio-temporal discretization.

\newtheorem{Claim}{Claim}
\begin{Claim}\label{C1}
    Events are triggered asynchronously, and under constant illumination, the same object moving at different speeds in the image plane will result in proportionally different local event densities.
\end{Claim}
\vspace{-0.2cm}

\begin{figure*}[t!]
    \begin{center}
        \vspace{-0.3cm}
        \centerline{\includegraphics[width=.85\linewidth]{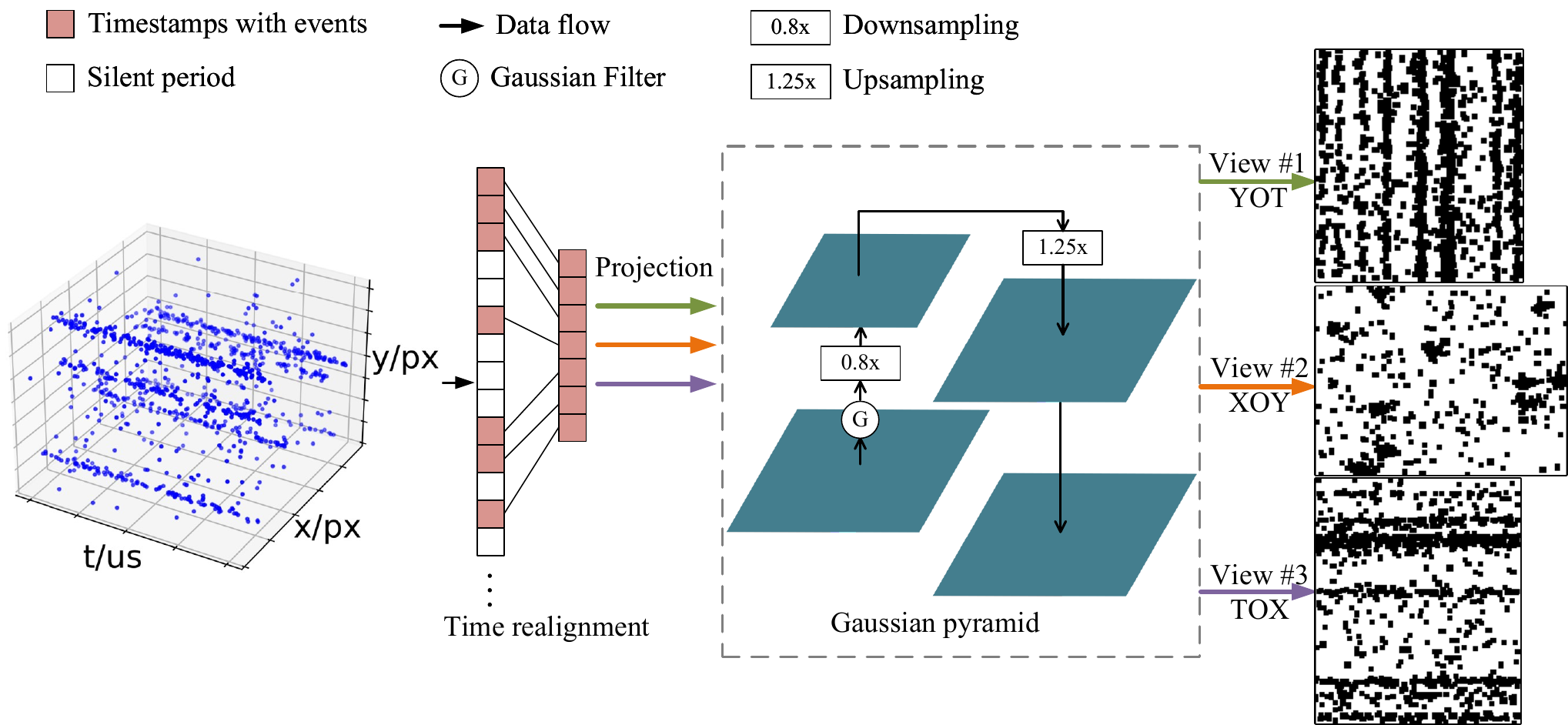}}
        \vspace{-0.3cm}
        \caption{Pipeline of the proposed MvP. The raw event data first undergoes time realignment to compress temporal gaps. The realigned events are then projected onto three orthogonal planes (YOT, XOY, and TOX). Finally, each view is refined through a Gaussian pyramid to enhance structural continuity, producing the final multi-view representation. The projections are grayscale, and we binarize them only for readability.}
        \label{fig4}
        \vspace{-0.7cm}
    \end{center}
\end{figure*}

\subsection{MvT Architecture}
\label{section:MvT}

MvT processes raw events (i.e. Dynamic Vision Sensor signals) and outputs event trajectories rather than bounding boxes. Input event streams are partitioned into non-overlapping batches based on a fixed event count or a fixed time window. Each batch is first encoded by MvP to generate the multi-view representation. Results obtained by image-based methods are then passed to MvF for complete trajectory recovery. Since views produced by MvP are independent, the image-based algorithm can be executed in parallel for efficiency.

\subsection{MvP}
\label{section:MvP}

MvP projects an event batch onto three orthogonal planes in the 3D Cartesian coordinate system, XOY, YOT, and TOX, as illustrated in Fig.~\ref{fig4}. Prior works in three-view reconstruction~\cite{kuo1998reconstruction} and spike-based coding~\cite{thorpeSpikebasedStrategiesRapid2001} demonstrate that those three views preserve sufficient information for recovering 3D structures. Despite reducing dimensionality to 2D, our representation retains essential event spatio-temporal information.

Ignoring polarities, each event is represented as $e=(t,x,y)$, where $t$, $x$, $y$ denote the raw timestamp and 2D locations, respectively. For a batch $\Phi=\{e_i=(t_i,x_i,y_i)|t_i\in[t_0,t_0+B]\}$ within a small window $[t_0,t_0+B]$, its timestamp set is $T=\{t_i|e_i\in \Phi\}$ and the range $\text{ran}(T)$ of $T$ is

\begin{equation}\label{eq_5}
    \text{ran}(T) = [{\rm min}(T),~{\rm max}(T)],
\end{equation}
where ${\rm min}(\cdot)$ denotes the minimum, and ${\rm max}(\cdot)$ denotes the maximum. Taking YOT as an example, the projected event count view is generated by:

\begin{equation}\label{eq_6}
    \mathrm{YOT}(y,~t) = \sum_{e_i\in \Phi}\delta (y-y_i,~t- t_i),
\end{equation}
where ${\mathrm{YOT}}\subset \mathbb{R}^{H\times |\text{ran}(T)|} = \mathbb{R}^{H\times (\mathrm {max}(T)-\mathrm {min}(T)+1)}$. $H$ denotes the height of spatial resolution. $|\cdot|$ represents the cardinality. $\delta(\cdot)$ is the Dirac delta function.

\textit{Time realignment.} To mitigate temporal discretization, we apply time realignment via a time warping function $g(\cdot)$

\begin{equation}\label{eq_7}
    \begin{aligned}
        \hat \Phi & = \{\hat e_i=(\hat t_i,x_i,y_i)|{\forall} \hat t_i \in \hat T\}, \\
        \hat T    & = g(T) \triangleq \{i|{\forall} t_i \in T, i\in \mathbb{N}\},
    \end{aligned}
\end{equation}
where $\hat \Phi$ is the time-realigned event set, and $\hat T$ is the set of realigned timestamps $\hat t$.

This process eliminates temporal gaps by renumbering timestamps sequentially from zero, as illustrated in Fig.~\ref{fig4}. By discarding redundant ``blank'' timestamps where no events are triggered, it compacts the timeline into a series of pseudo-timestamps. Therefore, $g(\cdot)$ is an inhomogeneous warping function that removes the gaps between timestamps. Since $g(\cdot)$ preserves order, it is also a monotonic function. This ensures that the mapping between $T$ and $\hat T$ is bijective, which is a prerequisite for timestamp recovery in the subsequent event trajectories.

The range of $\hat T$ is $\text{ran}(\hat T)$:

\begin{equation}\label{eq_8}
    \text{ran}(\hat T) = [0,~|T|-1].
\end{equation}

The realigned projection $\hat {\mathrm{YOT}} \subset \mathbb{R}^{H\times |\text{ran}( \hat T)|}$ achieves dimensional reduction since $|\text{ran}(\hat T)|=|T|\le |\text{ran}(T)|$. Owing to the simple implementation and monotonic characteristic, this adaptive time realignment can be seamlessly integrated into the voxel-grid-based representation for further temporal scale~\cite{zhang2021object}.

The quality of realigned timestamps is affected by two primary factors, multiple object velocities and noise that often refers to background activity (BA) events~\cite{guoLowCostLatency2023, jiangEDformerTransformerBasedEvent2024, duanLEDLargescaleRealworld2024}. Regarding the former, when objects exhibit varying planar velocities, the trajectories of slow-moving objects tend to be more fragmented and sparse along the temporal axis than those of fast-moving objects. However, the association model proposed in Sec.~\ref{section:MvF} can handle this fragmentation. With respect to the latter, BA noise introduces two distinct problems. First, it can distort the realigned timestamps away from their ideal (noise-free) values, leading to trajectory fragmentation. This issue is similarly addressed by the association model. Second, BA events may be wrongly detected as part of a tracklet. Given the inherently random and independent nature of BA events~\cite{chenDenoisingMethodDynamic2023}, even if a single noise event coincidentally overlaps with a tracklet in one 2D view, the probability of it consistently falling on the tracklet across all views is low. Consequently, the majority of BA events can be distinguished through cross-view validation.

\textit{Gaussian pyramid.}
Time realignment alleviates temporal sparsity, but discontinuities in tracklet patterns across the views persist due to spatial discretization and residual temporal sparsity. We employ a Gaussian pyramid consisting of dual pathways, bottom-up and top-down, to compensate for missing trajectory details where events are not triggered. By default, we set two scales for each pathway. This generates views with relatively continuous trajectory patterns, facilitating accurate detection and segmentation.

\begin{figure*}[t!]
    \begin{center}
        \vspace{-0.4cm}
        \centerline{\includegraphics[width=.9\linewidth]{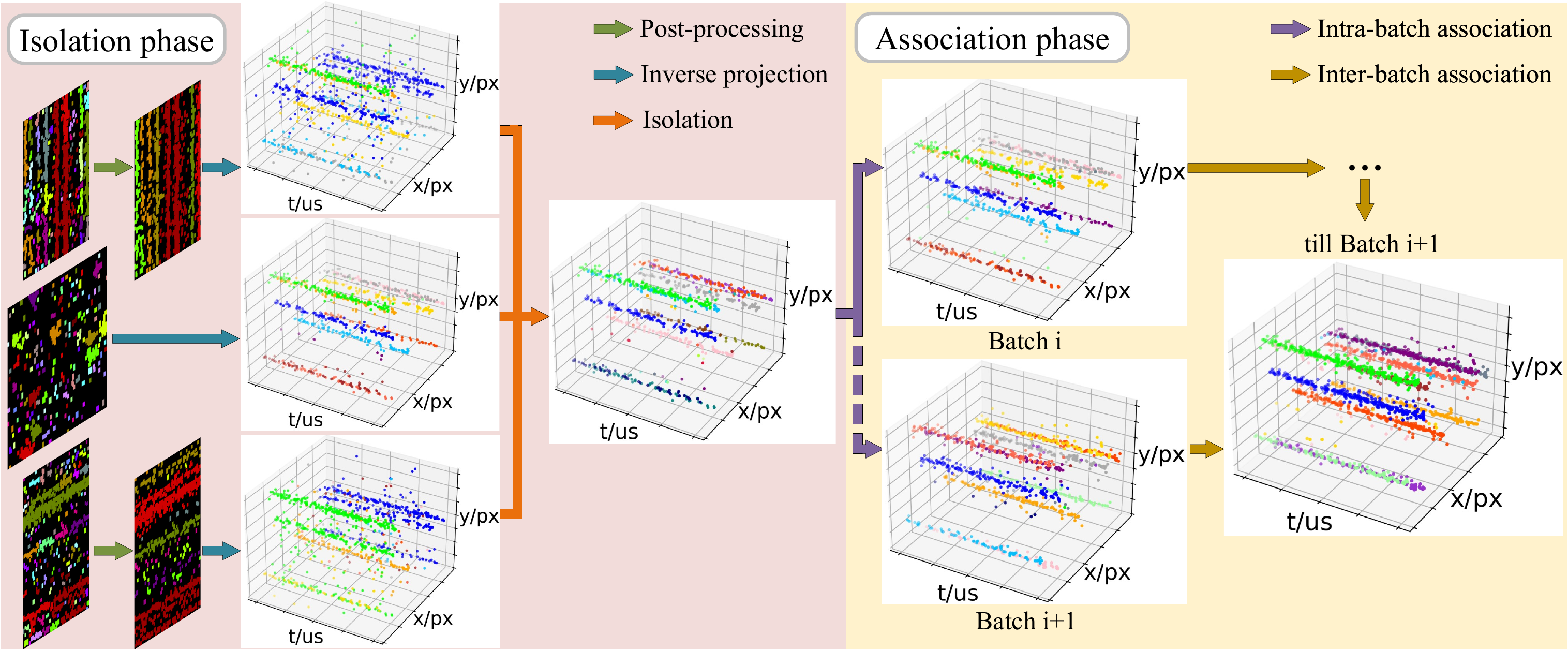}}
        \vspace{-0.2cm}
        \caption{Pipeline of the MvF stage. It consists of two core phases: (a) Isolation, which resolves trajectory ambiguities by integrating segmentation results from multiple views to extract distinct 3D trajectories, and (b) Association, which associates continuous and discontinuous tracklets into complete trajectories through optimal assignment. A simple filtering rule based on event count is applied to the final output to exclude isolated noise.}
        \label{fig5}
        \vspace{-0.4cm}
    \end{center}
\end{figure*}

The pyramid structure is shown in Fig.~\ref{fig4}. In the bottom-up pathway, Gaussian filtering with a kernel size of $k$ is applied prior to downsampling to prevent aliasing. The selection of $k$ and the corresponding standard deviation $\sigma$ are critical for effective anti-aliasing while mitigating pattern discontinuities. First, we determine the lower bound of $\sigma$ for anti-aliasing. According to the Nyquist sampling theorem, scaling by a factor of $s\in (0, 1)$, i.e., a sampling factor of $1/s$, reduces the Nyquist frequency from $f_N$ to $f'_N = s\cdot f_N$, where $f_N=1/2$ in a frame. To attenuate the signal to a level $\alpha$ at $f'_N$, the frequency response of the Gaussian function should satisfy $H(f'_N)=e^{-2\pi^2\sigma^2 {f'_N}^2}\le \alpha$, where $\alpha$ denotes the stopband edge gain and $\sigma$ is the standard deviation of the Gaussian kernel. Therefore, the lower bound of $\sigma$ is $\sigma_{anti} = \sqrt{\frac{-2\ln \alpha}{\pi^2s^2}}$. To determine the final $\sigma$ and $k$, we conduct a series of experiments to analyze the tracking performance under different combinations of $\sigma$ and $k$. The details are provided in Sec.~\ref{sec:gauss}. Although the spatial and temporal units are inhomogeneous, we still use isotropic Gaussian kernels. This is because our views, specifically YOT and TOX, are constructed by mapping physical timestamps onto discrete grid indices, which means that the Gaussian kernels operate on grid topology rather than physical timestamps. Moreover, our pyramid does not harm high-precision timestamps, since the smoothed views are used exclusively for detection/segmentation, whereas the other stages still retain the original timestamps.

BA events~\cite{wuProbabilisticUndirectedGraph2021, duanGuidedEventFiltering2021} always occur independently and randomly, unlike events from objects, which are highly correlated across neighboring pixels. Therefore, BA events are high-frequency anomalies superimposed on the ideal event output. When the Gaussian kernel is applied, neighboring pixels are averaged, which can suppress the influence of BA events. The top-down pathway upsamples the scaled frame. The pyramid design compensates for missing textures while incorporating an anti-aliasing kernel to minimize the impact of artificial artifacts. A detailed experimental analysis of this pyramid design is provided in Sec.~\ref{sec:AS}.

\subsection{MvF}
\label{section:MvF}
MvF is the core stage for recovering complete trajectories from multi-view detection or segmentation results. As illustrated in Fig.~\ref{fig5}, this stage operates in two sequential phases: (1) tracklet isolation, which integrates all results to generate isolated tracklets that are independent across views, and (2) tracklet association, which links tracklets for complete trajectories via optimal assignment.

\begin{figure}[t!]
    \begin{center}
        \vspace{-0.4cm}
        \centerline{\includegraphics[width=0.8\linewidth]{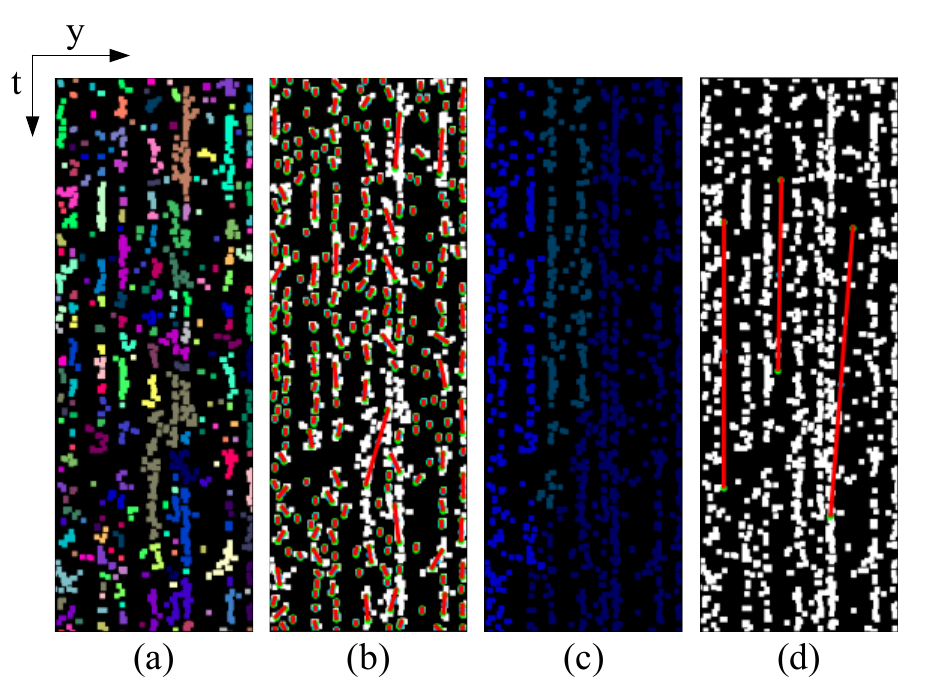}}
        \vspace{-0.4cm}
        \caption{Illustration of iterative region post-processing in the YOT view, where tracklets evolve along the t-axis. (a) Initial segmentation results, with colors denoting different detection or segmentation results. The tracklets are highly fragmented. (b) During post-processing, a direction vector $\mathbf{d}_i$ (red lines) is calculated from its two centroids $c_{i1}$ and $c_{i2}$ (green dots) for each tracklet fragment. (c) Merged tracklets after post-processing. (d) The final direction vectors and centroids recalculated for the tracklets are shown.}
        \label{fig6}
        \vspace{-0.7cm}
    \end{center}
\end{figure}

\textit{Tracklet isolation.}
The isolation phase leverages complementary information from different views to resolve trajectory ambiguities caused by occlusions. This phase is further divided into two steps, temporal post-processing and multi-view tracklet isolation. Due to the discretization of tracklet patterns along the t-axis or algorithmic errors, initial detection/segmentation results of YOT and TOX often consist of numerous small regions, as shown in Fig.~\ref{fig6}(a) and (b). Since sparse event points yield less accurate direction vectors in the association phase, we introduce an iterative post-processing step to combine regions, as shown in Fig.~\ref{fig6}(c) and (d).

\newtheorem{Criterion}{Criterion}
\begin{Criterion}\label{Cr1}
    Tracklets $K_i$ and $K_j$ of one trajectory should lie within the time-extended region of each other.
    \begin{equation}\label{eq_13}
        \Vert (c_j - c_i)\times \mathbf{d}_i\Vert_2 \leq w_i,
    \end{equation}
    where $c_i$, $\mathbf {d_i}$, and $w_i$ are the centroid, unit direction vector, and half-width of $K_i$. They are given by

    \begin{equation}\label{eq_9}
        c_i \triangleq (E[\hat T_i],~E[X_i],~E[Y_i]),
    \end{equation}
    \begin{equation}\label{eq_10}
        \mathbf d_i \triangleq \mathrm{normalize}(c_{i2}-c_{i1}),
    \end{equation}
    \begin{equation}\label{eq_11}
        w_i \triangleq \mathrm{max} (\Vert(K_i- c_i)\times \mathbf{d}_i\Vert_2),
    \end{equation}
    where $\hat T_i$, $X_i$, and $Y_i$ represent the realigned timestamp set and 2D location set of all events in $K_i$, respectively. $E[\cdot]$ means expectation. ${\mathrm {normalize}}(\cdot)$ denotes L2 normalization. $\Vert \cdot \Vert_2$ denotes L2-norm, and $c_{i1}$ and $c_{i2}$ are centroids of two parts of $K_i$ partitioned by its mean timestamp.
\end{Criterion}

Post-processing relies on \textbf{Criterion~\ref{Cr1}}, which is founded on a common assumption~\cite{zhu2017event, zihao2017event} that trajectories approximate linear tracklets within small spatio-temporal windows. Though this criterion describes relationships between 3D tracklets, it also applies to 2D segmentation regions.

After region post-processing, we perform inverse projection to map the multi-view results into a common 3D coordinate system. This step reconstructs the time-realigned 3D events from 2D segmentation regions $r_{XOY}$, $r_{YOT}$, and $r_{TOX}$, yielding corresponding 3D tracklet sets $R_{XOY}$, $R_{YOT}$, and $R_{TOX}$. Taking $R_{XOY}$ as an example, we formulate inverse projection in Eq.~\ref{eq_13_1}.

\vspace{-0.2cm}
\begin{small}
    \begin{equation}\label{eq_13_1}
        R_{XOY} = \{\hat e_i=(\hat t_i,x_i,y_i) \in \hat \Phi| \exists r_j  \in r_{XOY}, (x_i, y_i) \in r_j \},
    \end{equation}
\end{small}where $r_{XOY}$ consists of a set of regions $r_j$. Each region $r_j$ comprises a series of 2D coordinates.

The dimensionality reduction in the pseudo-frame representation means that events mapped to the same pixel may originate from distinct objects, which prevents single pseudo-frame methods from distinguishing intersecting or overlapping tracks.
Our solution for disambiguation is guided by the principle that distinct 3D trajectories can be separated as long as they are non-overlapping in at least one of their 2D projections. We isolate tracklets by intersecting their inverse-projected 3D event sets, leveraging the fact that overlapping tracks in one view are often separated in others.

\begin{figure}[t!]
    \begin{center}
        \vspace{-0.4cm}
        \centerline{\includegraphics[width=0.8\linewidth]{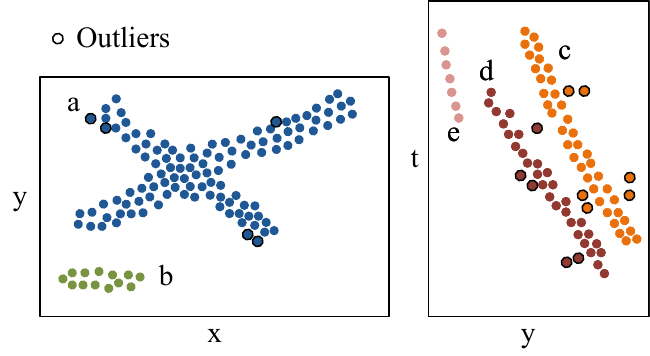}}
        \vspace{-0.4cm}
        \caption{Illustration of occlusion handling via multi-view trajectory isolation. The extracted trajectories from the XOY and YOT views are shown, where trajectory $a$ in the XOY corresponds to separated trajectories $c$ and $d$ in YOT, while trajectory $b$ corresponds to $e$. }
        \label{fig7}
        \vspace{-0.6cm}
    \end{center}
\end{figure}

To implement this, we systematically isolate tracklets via intersection operations on $R_{XOY}$, $R_{YOT}$, and $R_{TOX}$ to obtain the final isolated tracklet set $MR$, as in Eq.~\ref{eq_14}. This step also helps identify outliers within tracklets, as shown in Fig.~\ref{fig7}.

\begin{footnotesize}
    \begin{equation}\label{eq_14}
        MR = \{K_i\cap K_j\cap K_k|K_i\in R_{XOY}, K_j\in R_{YOT}, K_k\in R_{TOX}\}.
    \end{equation}
\end{footnotesize}

\textit{Tracklets association.}
The association phase handles two types of associations: (1) intra-batch association, resolving potential over-fragmentation due to isolation or inherent discontinuous trajectory patterns, and (2) inter-batch association, linking tracklets across different batches.
Although trajectories can be locally approximated as linear tracklets over short intervals, they often exhibit complex nonlinear patterns over extended periods due to the existence of varying forces, such as vehicle steering.
The inherent sparsity and asynchronicity of events further compound this challenge by producing non-smooth trajectory contours.

\begin{figure}[t!]
    \begin{center}
        \vspace{-0.3cm}
        \centerline{\includegraphics[width=0.95\linewidth]{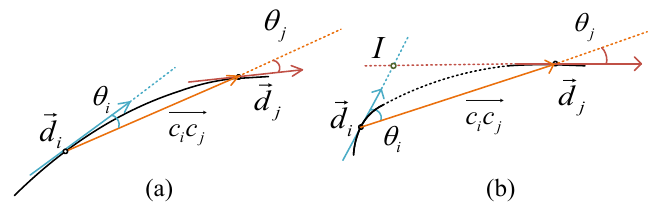}}
        \vspace{-0.3cm}
        \caption{Illustration of tracklet association criteria. $\theta_i$ and $\theta_j$ denote the angular deviations between the tracklet directions ($\mathbf{d}_i$, $\mathbf{d}_j$) and the normalized centroid connection vector $\mathbf{c}_i\mathbf{c}_j$, respectively. (a) \textbf{Criterion~\ref{Cr2}}: tracklets with small directional variation are associated through motion continuity constraints. (b) \textbf{Criterion~\ref{Cr3}}: tracklets exhibiting big directional changes are linked via spatio-temporal intersection verification.}
        \label{fig8}
        \vspace{-0.6cm}
    \end{center}
\end{figure}

Despite these complexities, adjacent tracklets typically maintain motion continuity. We formulate tracklet association as an optimal assignment problem governed by three geometric criteria, the previously introduced \textbf{Criterion~\ref{Cr1}}, along with \textbf{Criterion~\ref{Cr2}} and \textbf{\ref{Cr3}} illustrated in Fig.~\ref{fig8}. \textbf{Criterion~\ref{Cr2}} evaluates the coplanarity of two tracklets' directions and their positional variation, while \textbf{Criterion~\ref{Cr3}} assesses whether a valid spatio-temporal intersection $I$ exists between two tracklets.

\begin{Criterion}\label{Cr2}
    Tracklets $K_i$ and $K_j$ are considered to belong to the same trajectory if they achieve the minimal direction change among all candidate pairs, subject to constrained directional changes (not exceeding $\nicefrac{\pi}{2}$) and consistent positional progression.
    \begin{equation}\label{eq_15}
        \begin{aligned}
            \mathrm {argmin}_{K_i, K_j} & |(\mathbf{ d_i} \times \mathbf{ d_j})\cdot \mathbf{c}_i\mathbf{c}_j|,                                                         \\
            \mathrm {s.~t.}~~           & \Vert \mathbf{d}_i \times \mathbf {c_ic_j}\Vert_2 \le \beta,~~ \Vert \mathbf {c_ic_j} \times \mathbf{ d_j} \Vert_2 \le \beta,
        \end{aligned}
    \end{equation}where $\mathbf{c}_i\mathbf{c}_j$ is the normalized centroid connection vector, and $\beta$ is a parameter constraining the allowable directional change. If multiple pairs achieve the minimum, the tie is broken by selecting the pair with the smallest angle $\angle (\mathbf{d}_i, \mathbf{d}_j)$, and subsequently the smallest centroid distance.
\end{Criterion}

\begin{Criterion}\label{Cr3}
    For chronologically ordered tracklets $K_i$ and $K_j$ ($\hat t_{c_i} < \hat t_{c_j}$) that cannot satisfy \textbf{Criterion~\ref{Cr2}}, their tangent vectors are extended from centers $c_i$ and $c_j$ to compute an intersection point $I$. If $I$ lies within the spatio-temporal region bounded by their centers, they are likely to belong to one trajectory.
    \begin{small}
        \begin{equation}\label{eq_16}
            \begin{gathered}
                \exists ~ I \in S,                                                                                                             \\
                s.t.  \Vert \mathbf{d}_i \times \mathbf{d}_j\Vert_2 \ne 0,~~|(\mathbf{ d_i} \times \mathbf{ d_j})\cdot \mathbf{c}_i\mathbf{c}_j| = 0, \\
                S = \{(\hat t, x, y)|x \in [\min(x^{-}_{c_i},x^{-}_{c_j}), \max(x^{+}_{c_i},x^{+}_{c_j})], \\
                ~y \in [\min(y^{-}_{c_i},y^{-}_{c_j}), \max(y^{+}_{c_i},y^{+}_{c_j})],~\hat t \in [\hat t_{c_i},\hat t_{c_j}]\},
            \end{gathered}
        \end{equation}
    \end{small}where $x^{+}_{c_i}=x_{c_i}+w_i,~x^{-}_{c_i}=x_{c_i}-w_i,~y^{+}_{c_i}=y_{c_i}+w_i,~y^{-}_{c_i}=y_{c_i}-w_i$ define the spatial extents of x- and y-positions of $c_i$, $t_{c_i}$ is its realigned t-component, and $w_i$ denotes the spatial half-width of the spatio-temporal volume $K_i$. These variables for $c_j$ are defined analogously.
\end{Criterion}

To determine $\beta$ in \textbf{Criterion~\ref{Cr2}}, we first assume the ideal case where two tracklets from the same trajectory have a directional variation of $\theta_v \triangleq \angle (\mathbf{d}_i, \mathbf{d}_j) =\arcsin \Vert \mathbf{d}_i \times \mathbf{d}_j \Vert_2 \le \nicefrac{\pi}{2}$.
Let the angular deviation $\theta_i \triangleq \angle (\mathbf{d}_i, \mathbf{c}_i\mathbf{c}_j)$, $\theta_j \triangleq \angle (\mathbf{c}_i\mathbf{c}_j, \mathbf{d}_j)$. These three angles satisfy $\theta_i \le \theta_v$ and $\theta_j \le \theta_v$. The constraints in Eq.~\ref{eq_15} can be reformulated as

\begin{equation}\label{eq_17}
    \begin{aligned}
        \Vert \mathbf{d}_i \times \mathbf {c_ic_j} \Vert_2 \le \mathrm{sin}\theta_v, \\ \Vert \mathbf {c_ic_j} \times \mathbf{ d_j}\Vert_2 \le \mathrm{sin}\theta_v.
    \end{aligned}
\end{equation}

To account for computational errors in direction vectors, we introduce an angular error $\theta_e$ of $\theta_i$, $\theta_j$, and $\theta_v$, reformulating $\theta_v$ as $\theta_v + \theta_e$. Assuming $\theta_e \in [-\Delta \theta, \Delta \theta]$, the upper bound for $\beta$ is $\max(\sin(\theta_v \pm \Delta \theta))$. Given the computational error $\Delta \theta$, the candidate pairs with the minimal triple product in \textbf{Criterion~\ref{Cr2}} may no longer be the best matching pairs, requiring an adjustment to the objective. Since the original coplanarity objective minimizes the volume of the parallelepiped formed by unit vectors $\mathbf{d}_i$, $\mathbf{d}_j$, and $\mathbf{c}_i\mathbf{c}_j$, let $V=| (\mathbf{d}_i \times \mathbf{d}_j) \cdot \mathbf{c}_i\mathbf{c}_j |$ be the volume and $\mathbf{n}=\frac{\mathbf{d}_i \times \mathbf{d}_j}{\Vert \mathbf{d}_i \times \mathbf{d}_j \Vert_2}$ be the normal vector. Then

\begin{equation}\label{eq_18}
    \begin{aligned}
        V & = | \mathbf{d}_i \times \mathbf{d}_j | \cdot | \mathbf{c}_i\mathbf{c}_j | \\
          & = \sin \theta_v \cdot | \mathbf{n}\cdot \mathbf{c}_i\mathbf{c}_j |        \\
          & =\sin\theta_v | \cos \phi |   \le \sin \theta_v                           \\
          & \le \sin (\theta_i+\theta_j)\le \sin \theta_i + \sin \theta_j             \\
    \end{aligned}
\end{equation}
where $\phi$ is the angle between $\mathbf{n}$ and $\mathbf{c}_i\mathbf{c}_j$. The objective is simplified to minimizing the sum of the sines of these angular deviations, which aligns with the constraints.

\begin{equation}\label{eq_19}
    \begin{aligned}
        \mathrm {argmin}_{K_i, K_j} & \Vert \mathbf{ d_i} \times \mathbf {c_ic_j}\Vert_2 + \Vert \mathbf {c_ic_j} \times \mathbf{ d_j}\Vert_2 \\
        \mathrm {s.~t.}~~           & \Vert \mathbf{ d_i} \times \mathbf {c_ic_j}\Vert_2 \le \max(\sin(\theta_v\pm \Delta \theta)),           \\
                                    & \Vert \mathbf {c_ic_j} \times \mathbf{ d_j}\Vert_2 \le \max(\sin(\theta_v\pm \Delta \theta)).
    \end{aligned}
\end{equation}

Similarly, the triple product constraint in \textbf{Criterion~\ref{Cr3}} can be expressed with $\Delta \theta$. Consequently, this criterion is rewritten as Eq.~\ref{eq_21}, where $I$ is computed via vector operations and trigonometric relationships.

\vspace{-0.2cm}
\begin{small}
    \begin{equation}\label{eq_21}
        \begin{gathered}
            \exists ~ I \in S,            \\
            s.t.~ \Vert \mathbf{d}_i \times \mathbf{d}_j\Vert_2 \ne 0,\\
            |(\mathbf{ d_i} \times \mathbf{ d_j})\cdot \mathbf{c}_i\mathbf{c}_j| \le \sin (\theta_v\pm \Delta \theta), \\
            S = \{(\hat t, x, y)|x \in [\min(x^{-}_{c_i},x^{-}_{c_j}), \max(x^{+}_{c_i},x^{+}_{c_j})], \\
            ~y \in [\min(y^{-}_{c_i},y^{-}_{c_j}), \max(y^{+}_{c_i},y^{+}_{c_j})],~\hat t \in [\hat t_{c_i},\hat t_{c_j}]\},
        \end{gathered}
    \end{equation}
\end{small}

During intra-batch association, $MR$ is associated using \textbf{Criterion~\ref{Cr1}} and \textbf{\ref{Cr2}} sequentially, while in inter-batch association, \textbf{Criterion~\ref{Cr2}} and \textbf{\ref{Cr3}} are executed sequentially. It should be noted that \textbf{Criterion~\ref{Cr3}} has limited applicability for small objects. Since tracklets generated by small targets consist of only a few sparsely distributed events, their directional vectors are highly susceptible to local interference events. Due to geometric amplification, even a minor deviation in the estimated direction can cause the computed intersection $I$ to drift far outside the ideal spatio-temporal regions. Conversely, larger objects often trigger a large number of events, yielding more stable direction estimates. Furthermore, their larger volumes provide more tolerance for spatial deviation. This is confirmed by our ablation study in the supplementary material. Although the combination of these criteria is flexible, \textbf{Criterion~\ref{Cr3}} must be executed after \textbf{Criterion~\ref{Cr2}}. Specifically, it is applied only to the remaining unmatched tracklet pairs because it imposes a looser constraint, and applying this less restrictive condition as a fallback does not degrade tracking performance. Following the association process, successfully matched trajectories retain their original IDs, whereas newly detected trajectories are assigned new ones.

\begin{table*}
	\begin{center}
		\vspace{-0.3cm}
		\renewcommand\arraystretch{1}
		\caption{The configuration of SOD. ``Seq.'' means sequences, and ``Segments'' denotes the number of segments in each sequence.}
		\label{tab1}
		\vspace{-0.3cm}
		\begin{scriptsize}
			\begin{tabular}{
				>{\centering}p{0.8cm}|
				>{\centering}p{1cm}|
				>{\centering}p{1.1cm}|
				>{\centering}p{0.8cm}|
				>{\centering}p{0.7cm}|
				>{\centering}p{.9cm}|
				>{\centering}p{0.7cm}|
				>{\centering}p{1.3cm}|
				>{\centering}p{1.3cm}|
				>{\centering}p{1.4cm}|
				>{\centering}p{0.8cm}|
				>{\centering}p{0.6cm}|
				>{\centering\arraybackslash}p{1cm}
				}
				\Xhline{1.2pt}
				Category             & Source                    & Seq.      & Segments & Objects & Type      & Size                    & Scene        & Motion Type           & Speed (px/s)               & Camera                 & GT                 & Period (s) \\
				\Xhline{1.2pt}
				\multirow{3}*{SOD-r} & \multirow{3}*{Real-world} & Sphere-20 & 3        & 20      & Balls     & \multirow{6}{*}{5-10px} & Mixed        & Non-linear            & \multirow{3}*{$\ge$ 2500}  & \multirow{3}{*}{Fixed} & \multirow{3}*{No}  & 0.5        \\
				                     &                           & Sphere-30 & 3        & 30      & Balls     &                         & Mixed        & Non-linear            &                            &                        &                    & 0.5        \\
				                     &                           & Raindrops & 3        & $>$ 100 & Raindrops &                         & Rain         & Linear                &                            &                        &                    & 4.0        \\
				\cline{1-6}\cline{8-13}
				\multirow{3}*{SOD-b} & \multirow{3}*{Blender}    & Tree      & 1        & 19      & Balls     &                         & Tree         & \multirow{3}{*}{Free} & \multirow{3}{*}{$\ge$ 100} & \multirow{3}{*}{6DoF}  & \multirow{3}*{Yes} & 3.0        \\
				                     &                           & Sky       & 1        & 28      & Balls     &                         & Cloudy sky   &                       &                            &                        &                    & 3.0        \\
				                     &                           & Smog      & 1        & 44      & Balls     &                         & Full of smog &                       &                            &                        &                    & 3.0        \\
				\Xhline{1.2pt}
			\end{tabular}
		\end{scriptsize}
		\vspace{-0.6cm}
	\end{center}
\end{table*}

\section{Experiments}
\label{section:exps}

Our experiments have four purposes. First, assess the generalization capability and tracking effectiveness of the proposed framework MvT by incorporating both non-learning and learning-based image-based detection and segmentation methods.
Second, compare MvT with existing pseudo-frame and event-by-event-based baselines, with a particular focus on small object tracking scenarios. Third, perform ablation studies to validate the contribution of components in MvT. Finally, sensitivity experiments on key parameters, including direction angle error $\Delta \theta$, sampling scale $s$ in Gaussian pyramid, and window size $B$, are provided in the supplementary material. All evaluations are conducted on a platform with an Intel Core i7-7700HQ CPU @ 2.8GHz, 16G pf 2600MHz RAM and an NVIDIA GTX 1050 Ti GPU. Since event cameras suffer from significant noise~\cite{ding2023mlb}, we apply an event count filter to all methods' results to remove trajectories with fewer than $m$ events. $m$ is set to 20, which is lower than the total event count of a legitimate and complete trajectory.

\subsection{Datasets}

\begin{figure}[t!]
	\begin{center}
		\centerline{\includegraphics[width=1\linewidth]{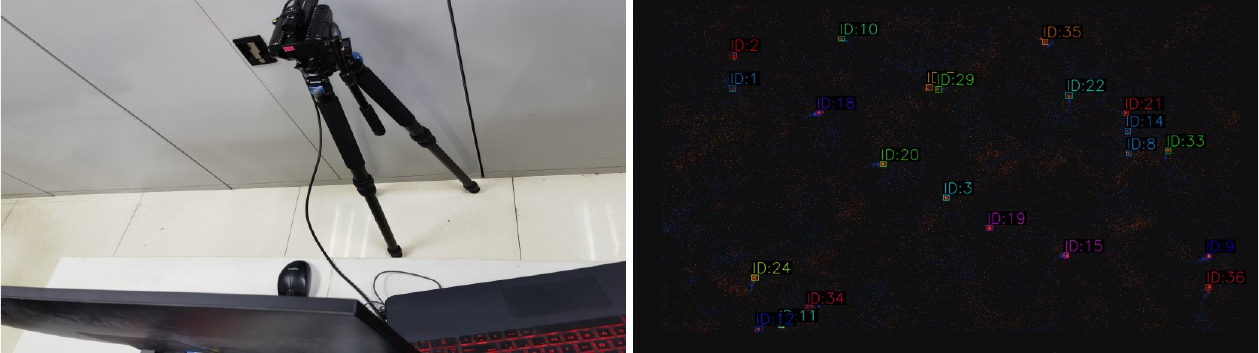}}
		\vspace{-0.2cm}
		\caption{The recording equipment and smog scenario setup in the XOY view used in our SOD-b dataset.}
		\label{fig_sod_b}
		\vspace{-0.6cm}
	\end{center}
\end{figure}

\label{section:Data}
\textit{Datasets.}
Experiments are conducted on four datasets with varying object scales, including three public datasets, ECD Shapes~\cite{mueggler2017event}, BEES~\cite{apps2025asynchronous}, and Ev-UAV~\cite{chenEventbasedTinyObject2025}, and a self-collected dataset SOD. ECD is a standard bounding box-based MOT dataset, containing different shapes of objects with normal size ($>25\times 25$ pixels) under three camera motions, translation, rotation, and 6DoF. BEES and Ev-UAV provide event-wise Ground Truth (GT) labels and focus on small objects ($\le 10 \times 10$ pixels) with complex and independent motion patterns. BEES, despite being a 5$s$ recording, contains approximately 5 million events with 70 to 110 labeled bees. However, it only provides segmentation labels without distinguishing different trajectories. Ev-UAV includes both segmentation and tracking labels, but only a few UAVs ($\le$ 5) appear in the test and validation sets.

Due to the limited number of datasets for small and multiple object tracking in event-based vision, we collect SOD to include small, independently fast-moving targets for evaluation. Unlike the previous three datasets with low spatial resolution ($240 \times 360$) and a single moving element (either the camera or the objects), SOD is recorded using a high spatial resolution ($720 \times 1280$) event camera, Prophesee EVK4, and provides scenarios with both camera and object motion. It consists of two parts, real-world and simulated data, as listed in Table~\ref{tab1}. The real-world sequences are captured in three outdoor scenarios, tree, playground, and rain. Based on the target types and quantities, these are categorized as Sphere-20, Sphere-30, and Raindrops. Sphere-20 and Sphere-30 sequences record marbles thrown by hand in both tree and clean playground scenes, while the Raindrops sequences record falling raindrops. The simulated sequences are generated by screen-recording the scenarios rendered by Blender, as shown in Fig.~\ref{fig_sod_b}, which is a common practice in event camera simulation~\cite{chin2019star}. In the simulation set, we design three different scenarios, tree, sky and smog, within which we simulate blobs moving randomly. Using the bounding box-based tracking GTs generated by Blender, we convert them into event-wise labels following the same procedure as in Ev-UAV. While Ev-UAV mainly contains trajectories with large spatio-temporal breaks, SOD features target overlapping and static background occlusion. SOD will be made available in our repository, and more information about the simulated subset (SOD-b) is provided in the supplementary material.

\begin{figure*}[t!]
	\begin{center}
		\vspace{-0.3cm}
		\centerline{\includegraphics[width=.9\linewidth]{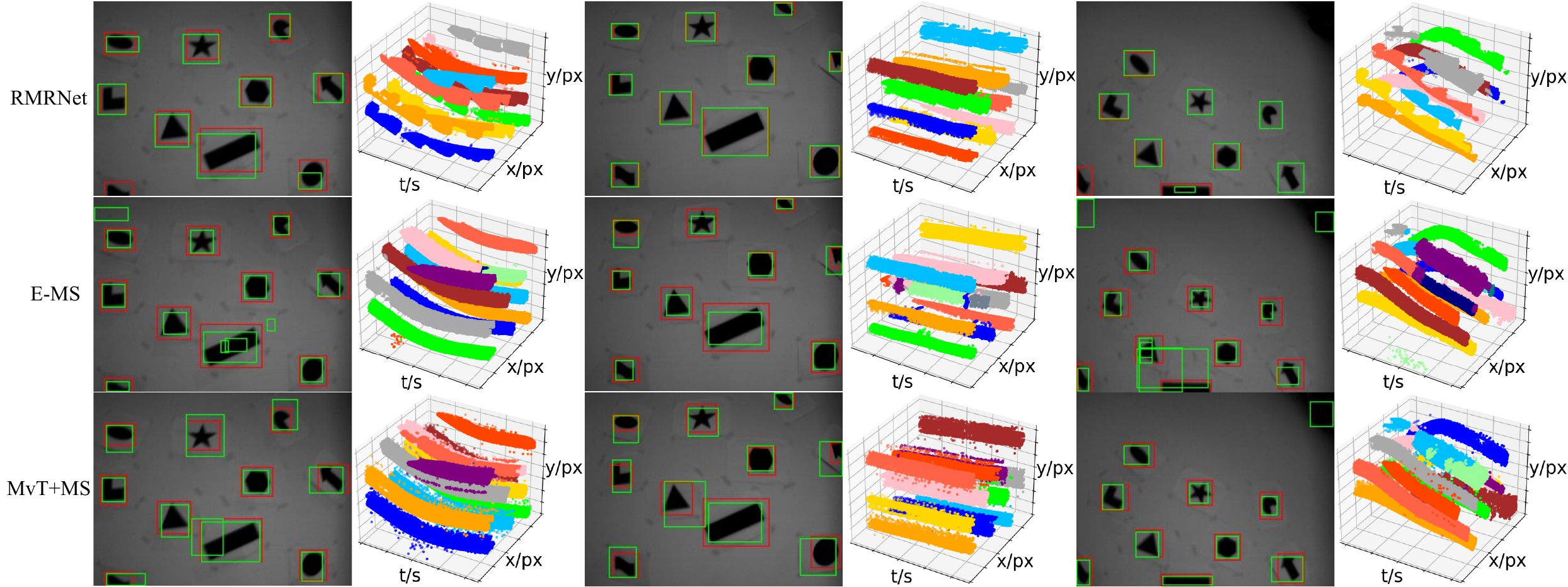}}
		\vspace{-0.2cm}
		\caption{Partial tracking results on ECD Shapes~\cite{mueggler2017event} are presented in 2D and 3D. In 2D images (the XOY view), red boxes denote GT, while green boxes represent the tracking results. For 3D visualizations, the displayed period is 200 \textit{ms} around the timestamp of the left image. Distinct colors represent individual trajectories. From left to right, the sequences are translation (frame 532), 6DoF (frame 236), and 6DoF (frame 1014).}
		\label{fig9}
		\vspace{-0.6cm}
	\end{center}
\end{figure*}

\subsection{Evaluation metrics}
To comprehensively evaluate our framework, we employ both bounding box-based and event-wise MOT metrics. The former aligns with standard metrics in image-based MOT, including Precision (P), Recall (R), Multiple Object Tracking Accuracy (MOTA)~\cite{luitenHOTAHigherOrder2021}, and Feature Age (FA)~\cite{alzugaray2020haste}. By default, FA is evaluated over all trajectories. For a fair comparison, we evaluate only the longest trajectories, with counts equalized to 20\% and 50\% of the minimum trajectory total across all methods, visualized via box plots. In these plots, boxes represent the interquartile range (25\%-75\%), with blue and red lines marking the median and mean, and whiskers showing the full data range. Event-wise metrics are derived from event-based segmentation metrics~\cite{chenEventbasedTinyObject2025}, specifically Segmentation IoU (Seg. IoU) and Segmentation Accuracy (Seg. Acc). We extend them to Track IoU and Track Acc, and add IDF1~\cite{luitenHOTAHigherOrder2021} to evaluate tracking stability. As box-based metrics rely on GT bounding boxes, they may yield biased evaluations when events of one trajectory are temporarily misassigned to another trajectory. Therefore, event-wise metrics serve as the primary indicators.

\subsection{Integration with non-learning-based methods}
We integrate three classic image-based segmentation methods, region growing~\cite{hojjatoleslami1998region}, watershed~\cite{vincent1991watersheds}, and Mean Shift~\cite{comaniciu2003kernel} into MvT, denoted as MvT+RG, MvT+WS, and MvT+MS. Implementations are provided in the supplementary material.

\textit{Comparing Methods.}
\label{section:SOTA}
In this section, we compare our MvT with existing event-based MOT methods. For pseudo-frame methods, we compare against ETD~\cite{chen2019asynchronous} and RMRNet~\cite{chen2020end}. For the event-by-event paradigm, we evaluate against AEBTracker~\cite{2024_Wang_AEB_Tracker_TRO} (AEBT), AEMOT~\cite{apps2025asynchronous}, and E-MS~\cite{barranco2018real}. Since ETD, RMRNet, and AEBT require GT initialization of object positions, they are evaluated exclusively on ECD. AEMOT incorporates a binary classifier trained on $28 \times 28$ intensity patches, which are only available in BEES. For BEES and SOD, we additionally employ the spatter tracking method from Prophesee software~\cite{prophesee2021tracking} as a baseline, denoted as PROPHESEE. Since PROPHESEE generates bounding boxes at each evaluation window, we convert the boxes into event trajectories for evaluation and visualization.

\textit{(1) Tracking on ECD Shapes.}
\label{section:MOT_task}
All compared methods are evaluated using their original settings. RMRNet is trained on the 6DoF sequence and tested on the remaining scenes. 
For MvT, we set $B=1800$ events (consistent with E-MS) and $\Delta \theta=\pi/6$.
Although ECD provides only bounding box-based GTs, we convert them into event-level GT trajectories using the same rule as in Ev-UAV to enable event-wise evaluation. Results are presented in Table~\ref{tab2} and Fig.~\ref{fig9}. In Table~\ref{tab2}, methods requiring GT initialization always rank in the top two for Precision. This advantage is attributed not only to GT initialization but also to the upstream learning-based or appearance-based detectors. Event cameras are inherently sensitive to dynamic changes, and in ECD, environmental artifacts such as paper contours or wall shadows frequently trigger events. In non-learning-based methods, these events may be mistakenly identified as objects, increasing false positives and reducing overall precision.


\begin{table}
	\begin{center}
		\renewcommand\arraystretch{1}
		\vspace{-0.2cm}
		\setlength{\abovecaptionskip}{-0.15cm}
		\caption{Results of MOT on ECD Shapes. Here ``GTs'' indicates whether the method uses GTs initialization. ``trans.'' and ``rotat.'' are translation and rotation sequences. ``FPS'' refers to the processing batches/frames per second. The \textbf{best} and \underline{second best} values are highlighted.}
		\label{tab2}
		\begin{scriptsize}
			\hspace*{-0.25cm}
			\begin{tabular}{
				>{\centering}p{0.35cm}|
				>{\centering}p{0.3cm}|
				>{\centering}p{0.75cm}|
				>{\centering}p{0.45cm}|
				>{\centering}p{0.45cm}|
				>{\centering}p{0.64cm}|
				>{\centering}p{0.65cm}|
				>{\centering}p{0.65cm}|
				>{\centering}p{0.45cm}|
				>{\centering\arraybackslash}p{1.0cm}
				}
				\Xhline{1.2pt}
				\multirow{2}*{Seq.}   & \multirow{2}*{GTs} & \multirow{2}*{Method} & P                 & R                 & MOTA              & Track             & Track             & IDF1              & \multirow{2}*{FPS}    \\
				                      &                    &                       & (\%)              & (\%)              & (\%)              & IoU(\%)           & Acc(\%)           & (\%)              &                       \\
				\Xhline{1.2pt}
				\multirow{7}*{trans.} & \multirow{3}*{Yes} & ETD                   & 48.72             & 40.61             & -6.31             & --                & --                & --                & 7.44$\times$10$^{-7}$ \\

				~                     & ~                  & RMRNet                & \textbf{95.88}    & 91.69             & --                & --                & --                & --                & 6.67                  \\

				~                     & ~                  & AEBT                  & \underline{81.87} & 81.73             & 54.11             & --                & --                & --                & 2.51                  \\
				\cline{2-10}
				~                     & \multirow{4}*{No}  & E-MS                  & 74.74             & 97.24             & 62.56             & 30.51             & 40.57             & 40.36             & 2.81$\times$10$^{4}$  \\

				~                     & ~                  & \textit{MvT+RG}       & {75.85}           & \textbf{99.36}    & \underline{64.67} & \textbf{43.73}    & \textbf{55.69}    & 72.84             & 2.58$\times$10$^{1}$  \\

				~                     & ~                  & \textit{MvT+WS}       & {77.92}           & 97.61             & \textbf{65.49}    & 26.93             & 30.68             & \underline{74.64} & 2.16$\times$10$^{1}$  \\

				~                     & ~                  & \textit{MvT+MS}       & 76.49             & \underline{98.91} & 64.49             & \underline{43.43} & \underline{54.46} & \textbf{75.87}    & 1.8$\times$10$^{-1}$  \\
				\hline
				\multirow{7}*{rotat.} & \multirow{3}*{Yes} & ETD                   & 41.06             & 48.04             & -26.63            & --                & --                & --                & 7.78$\times$10$^{-7}$ \\

				~                     & ~                  & RMRNet                & \textbf{93.72}    & 88.00             & --                & --                & --                & --                & 5.83                  \\

				~                     & ~                  & AEBT                  & {69.19}           & 86.01             & 31.98             & --                & --                & --                & 2.51                  \\
				\cline{2-10}
				~                     & \multirow{4}*{No}  & E-MS                  & 65.64             & 95.86             & 43.11             & 29.37             & 39.77             & 57.68             & 2.67$\times$10$^{4}$  \\

				~                     & ~                  & \textit{MvT+RG}       & {65.73}           & \textbf{99.11}    & {42.47}           & \underline{37.46} & \textbf{48.81}    & 65.97             & 2.43$\times$10$^{1}$  \\

				~                     & ~                  & \textit{MvT+WS}       & {69.72}           & 97.42             & \underline{47.80} & 23.70             & 24.56             & \underline{66.33} & 1.93$\times$10$^{1}$  \\

				~                     & ~                  & \textit{MvT+MS}       & \underline{69.89} & \underline{98.45} & \textbf{49.56}    & \textbf{37.78}    & \underline{46.33} & \textbf{70.98}    & 1.8$\times$10$^{-1}$  \\
				\hline
				\multirow{7}*{6dof}   & \multirow{3}*{Yes} & ETD                   & 48.87             & 34.61             & -3.62             & --                &                   &                   & 7.23$\times$10$^{-7}$ \\

				~                     & ~                  & RMRNet                & \textbf{99.72}    & 97.18             & --                & --                & --                & --                & 7.78                  \\

				~                     & ~                  & AEBT                  & \underline{83.38} & 72.01             & 48.76             & --                & --                & --                & 3.72                  \\
				\cline{2-10}
				~                     & \multirow{4}*{No}  & E-MS                  & 67.93             & 97.29             & 43.55             & 23.29             & 21.80             & 24.60             & 2.59$\times$10$^{4}$  \\

				~                     & ~                  & \textit{MvT+RG}       & {77.90}           & \textbf{99.48}    & \underline{67.83} & \textbf{32.54}    & \textbf{37.05}    & \underline{57.21} & 2.45$\times$10$^{1}$  \\

				~                     & ~                  & \textit{MvT+WS}       & 80.07             & 97.42             & \textbf{68.56}    & {19.35}           & 18.28             & 56.33             & 1.90$\times$10$^{1}$  \\

				~                     & ~                  & \textit{MvT+MS}       & 77.18             & \underline{98.80} & 64.73             & \underline{30.01} & \underline{36.33} & \textbf{57.94}    & 1.9$\times$10$^{-1}$  \\
				\Xhline{1.2pt}
			\end{tabular}
		\end{scriptsize}
		\vspace{-0.6cm}
	\end{center}
\end{table}


\begin{table}
	\begin{center}
		\renewcommand\arraystretch{1}
		\vspace{-0.cm}
		\setlength{\abovecaptionskip}{-0.15cm}
		\caption{Tracking performance on BEES.}
		\vspace{0.3cm}
		\label{tab2_5}
		\begin{tabular}{
			>{\centering}p{1.6cm}|
			>{\centering}p{1.5cm}|
			>{\centering}p{1.5cm}|
			>{\centering\arraybackslash}p{2.2cm}
			}
			\Xhline{1.2pt}
			Method          & Seg. IoU(\%)      & Seg. Acc(\%)      & Number of Tracks \\
			\Xhline{1.2pt}
			AEMOT           & 36.73             & 37.83             & 297              \\
			E-MS            & 29.52             & \textbf{98.08}    & 17686            \\
			PROPHESEE       & 28.39             & 33.99             & 954              \\
			\hline
			\textit{MvT+RG} & \textbf{54.81}    & {57.76}           & 351              \\
			\textit{MvT+WS} & {29.93}           & 30.42             & 276              \\
			\textit{MvT+MS} & \underline{53.44} & \underline{71.66} & 630              \\

			\Xhline{1.2pt}
		\end{tabular}
		\vspace{-0.7cm}
	\end{center}
\end{table}

In terms of the tracking metrics MOTA and IDF1, the MvT variants (MvTs) achieve the top three across all scenarios. Unlike Track IoU and Track Acc, which only reflect the percentage of true positive events within each GT trajectory, the leading MOTA and IDF1 scores of MvTs demonstrate stronger long-term tracking and association capabilities compared to other methods. As shown in Fig.~\ref{fig9}, RMRNet produces less smooth trajectories than other methods, as it treats event-based tracking purely as an image-based task. MvT+MS, which utilizes the same clustering strategy Mean Shift as E-MS, yields more accurate tracking and continuous trajectories than E-MS. Among MvTs, MvT+RG and MvT+MS achieve the best overall performance, while MvT+WS consistently yields the lowest Track IoU and Track Acc, revealing that it fails to capture the majority of target events.

The runtime efficiency of MvT depends primarily on the adopted image-based method. As shown in Table~\ref{tab2}, MvT+RG and MvT+WS achieve near real-time speeds of 19-26 FPS, which is higher than RMRNet and ETD. MvT+MS drops to 0.18-0.19 FPS because the Mean Shift in the scikit-learn module is computationally expensive and cannot be executed in parallel. Nevertheless, MvT strikes a balance between efficiency and tracking accuracy.

\begin{table*}[t]
	\begin{center}
		\renewcommand\arraystretch{1}
		\setlength{\abovecaptionskip}{0.1cm}
		\vspace{-0.4cm}
		\caption{Results on Ev-UAV. ``*'' means the image-based method. ``Train'' column indicates whether the method needs to be trained. ``Rep.'' means the representation. In ``Event Rep.'' column, ``Event Count'' is event count images, one of the pseudo-frame representation, while ``Point'' is the original event data format. Except for two SOT models, the FPS for the remaining models is computed excluding the detection/segmentation time (or XOY processing time for MvT).} 
		\label{tab8}
		\vspace{-0.4cm}
		\hspace*{-0.25cm}
		\begin{tabular}{
			>{\centering}p{1.6cm}|
			>{\centering}p{0.9cm}|
			>{\centering}p{1.2cm}|
			>{\centering}p{0.6cm}|
			>{\centering}p{1.5cm}|
			>{\centering}p{1.5cm}|
			>{\centering}p{1.5cm}|
			>{\centering}p{1.6cm}|
			>{\centering}p{1.7cm}|
			>{\centering}p{.9cm}|
			>{\centering\arraybackslash}p{0.7cm}
			}
			\Xhline{1.2pt}
			Method      & Task     & Source    & Train & Event Rep.  & Seg. IoU(\%)   & Seg. Acc(\%)   & Track IoU(\%)  & Track Acc(\%)  & IDF1(\%)       & FPS             \\
			\Xhline{1.2pt}
			HDETrack    & SOT      & 2024CVPR  & Yes   & Event Count & \textbf{74.06} & \textbf{86.87} & \textbf{72.86} & \textbf{86.02} & \textbf{99.00} & 8.66            \\
			+MvT        & MOT      & --        & No    & MvP         & 65.29          & 72.71          & 60.91          & 67.17          & 73.15          & \textbf{147.68} \\
			\hline
			TSETrack    & SOT      & 2026IJCV  & Yes   & Event Count & \textbf{78.03} & \textbf{88.55} & \textbf{76.00} & \textbf{88.22} & \textbf{99.50} & 13.98           \\
			+MvT        & MOT      & --        & No    & MvP         & 68.88          & 74.42          & 60.17          & 67.49          & 75.14          & \textbf{146.01} \\
			\hline
			YOLOX*      & Detect.  & 2021ArXiv & Yes   & Event Count & \textbf{69.58} & \textbf{70.86} & --             & --             & --             & --              \\
			+OCSORT*    & MOT      & 2023CVPR  & No    & Event Count & 61.25          & 62.02          & 35.70          & 29.44          & 39.72          & \textbf{615.75} \\
			+MvT        & MOT      & --        & No    & MvP         & 59.77          & 60.48          & \textbf{50.79} & \textbf{54.44} & \textbf{71.74} & {170.93}        \\
			\hline
			SEGMAN*     & Segment. & 2025CVPR  & Yes   & Event Count & \textbf{57.65} & \textbf{58.34} & --             & --             & --             & --              \\
			+DEVA*      & MOT      & 2023ICCV  & Yes   & Event Count & 18.49          & 18.53          & 20.35          & 15.07          & 26.72          & 3.41            \\
			+MvT        & MOT      & --        & No    & MvP         & 48.62          & 49.10          & \textbf{34.67} & \textbf{43.73} & \textbf{70.41} & \textbf{19.10}  \\
			\hline
			RVT         & Detect.  & 2023CVPR  & Yes   & Voxel-grid  & \textbf{18.88} & \textbf{19.96} & --             & --             & --             & --              \\
			+OCSORT*    & MOT      & 2023CVPR  & No    & MvP         & 14.52          & 14.90          & 18.61          & 6.18           & 23.78          & 72.98           \\
			+MvT        & MOT      & --        & No    & MvP         & 15.81          & 16.36          & \textbf{33.04} & \textbf{13.99} & \textbf{39.22} & \textbf{214.75} \\
			\hline
			EV-SpSegNet & Segment. & 2025ICCV  & Yes   & Point       & 42.67          & {65.78}        & --             & --             & --             & --              \\
			+MvT        & MOT      & --        & No    & MvP         & \textbf{45.16} & \textbf{74.31} & \textbf{60.26} & \textbf{64.95} & \textbf{64.69} & \textbf{22.44}  \\
			\Xhline{1.2pt}
		\end{tabular}
		\vspace{-0.6cm}
	\end{center}
\end{table*}

\begin{figure}[t!]
	\begin{center}
		\vspace{-0.1cm}
		\centerline{\includegraphics[width=1\linewidth]{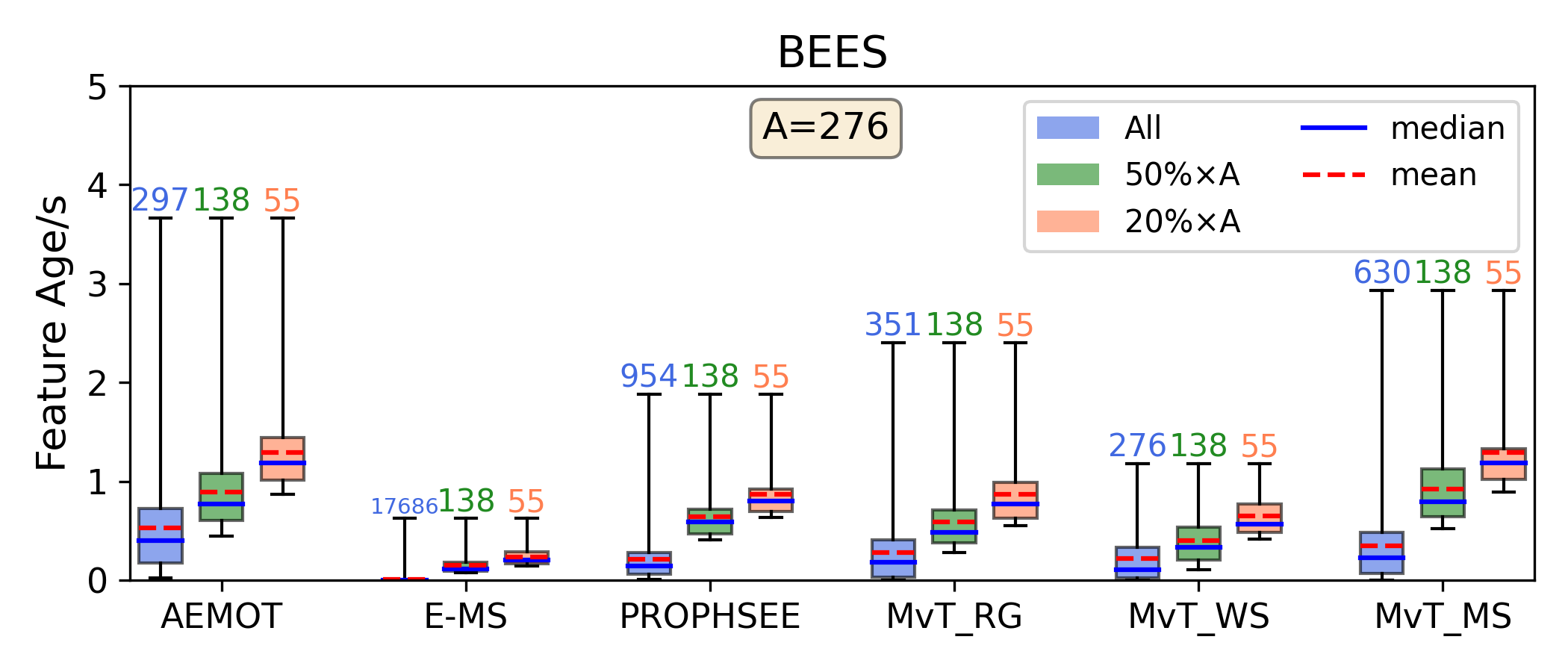}}
		\vspace{-0.4cm}
		\caption{Box plots of Feature Age on BEES. }
		\label{fig11}
		\vspace{-1.cm}
	\end{center}
\end{figure}

\textit{(2) Tracking on Bee Swarm Dataset.}
We set $B=5$ \textit{ms} according to the evaluation protocol in the BEES dataset~\cite{apps2025asynchronous} for all methods. The configuration of MvT remains unchanged. The bandwidth of E-MS is set to 0.1, and AEMOT follows its original settings. Table~\ref{tab2_5} primarily reports segmentation performance, while tracking performance is reflected by the FA metric in Fig.~\ref{fig11}. Regarding the two segmentation metrics, MvT variants rank first and second in Seg. IoU, and second and third in Seg. Acc. Since Seg. Acc is equivalent to the detection metric Recall, it only reflects the proportion of target events captured. As shown in Table~\ref{tab2_5}, E-MS achieves 98\% Acc but a low IoU, indicating it captures nearly all target events at the cost of numerous false positives. In terms of FA, AEMOT achieves the longest trajectory lifetime, while other four methods except E-MS yield comparable mean and median values. As illustrated in Fig.~\ref{fig10}, AEMOT also yields the cleanest event stream output, which is attributed to its learning-based detector's ability to effectively distinguish targets from background noise. However, it misses trajectories in certain regions (highlighted in two colored boxes) that other methods capture. Among the remaining methods, MvT+RG outperforms PROPHESEE by generating a greater number of clear trajectories under noisy conditions (blue boxes).

Note that FA values in Fig.~\ref{fig11} exhibit an inverse correlation with the number of tracked objects. This is because, under a fixed number of targets and a fixed total duration, a higher trajectory count inherently implies that continuous trajectories are being fragmented regardless of misdetections. The FA comparisons for the Top-20\% and Top-50\% longest trajectories show the same trend as the overall FA. These results collectively confirm that our MvT framework is effective for tracking small objects with complex motion patterns.

\begin{figure}[t!]
	\begin{center}
		\centerline{\includegraphics[width=1\linewidth]{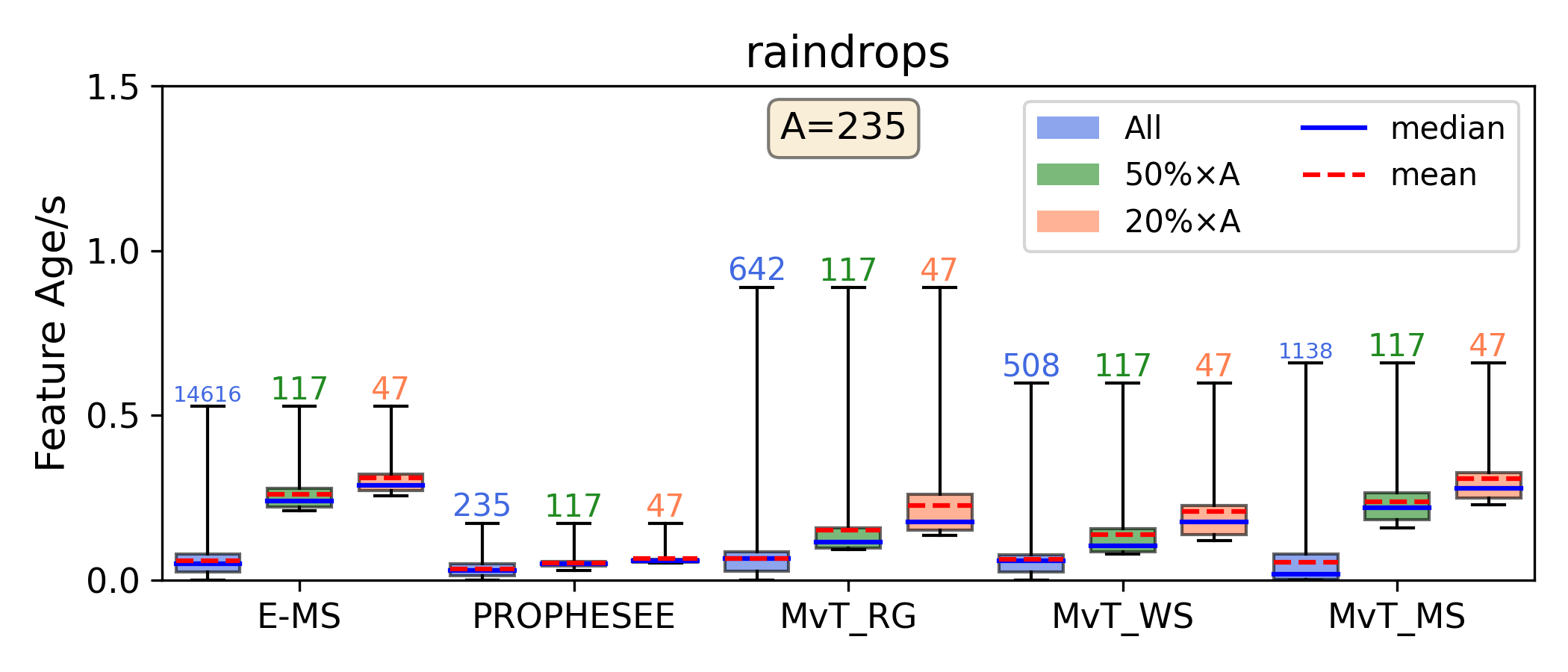}}
		\vspace{-0.4cm}
		\caption{Box plots of Feature Age on SOD-r Raindrops.}
		\label{fig13}
		\vspace{-1.cm}
	\end{center}
\end{figure}

\begin{figure*}[t!]
	\begin{center}
		\vspace{-0.4cm}
		\centerline{\includegraphics[width=0.95\linewidth]{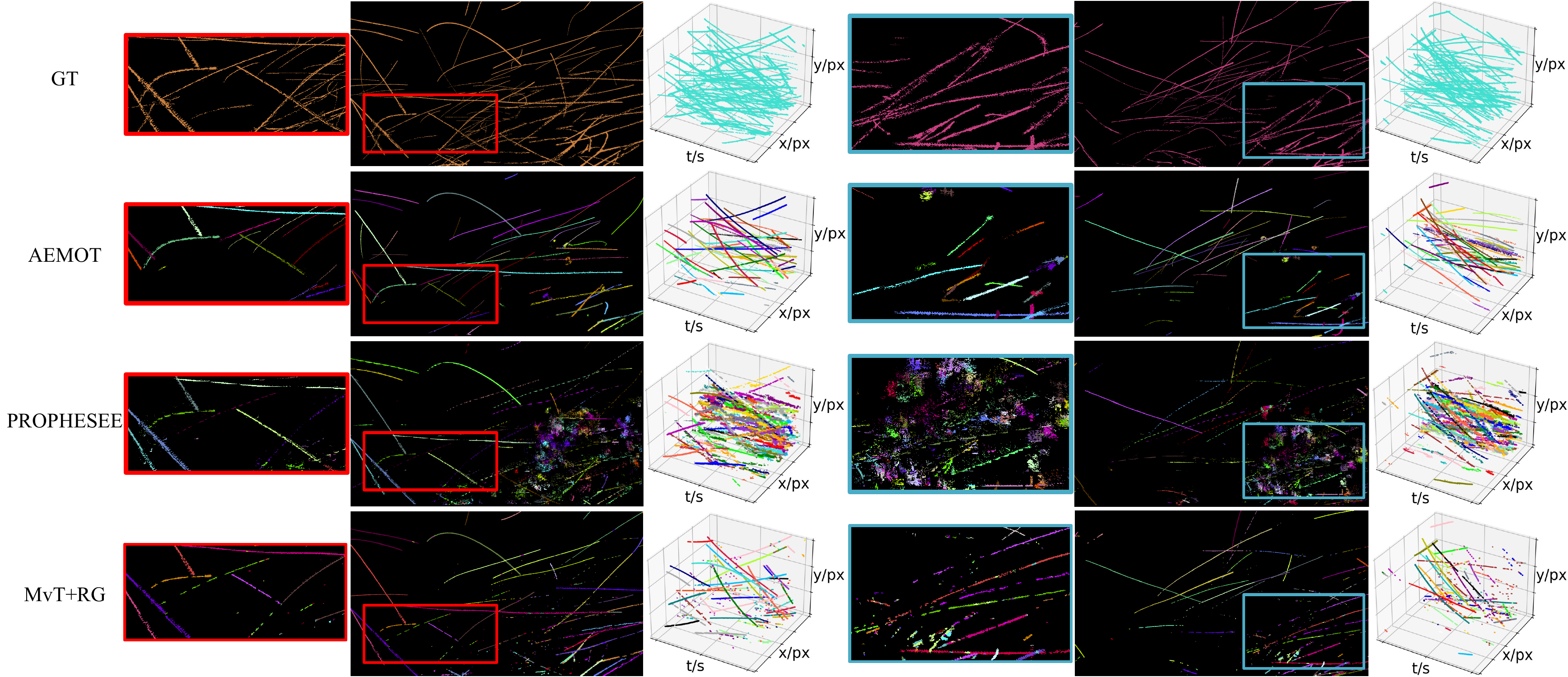}}
		\vspace{-0.2cm}
		\caption{Tracking result examples on the BEES dataset for GT, AEMOT, Prophesee SDK, and MvT+RG, visualized across two periods (2-2.5 $s$ and 3.5-4 $s$). 2D images (XOY views) are synthesized by accumulating all tracked event trajectories, where black pixels indicate no events and other colors correspond to distinct trajectories. Note that official GTs do not distinguish between individual bee trajectories, so they are shown in a uniform color.}
		\label{fig10}
		\vspace{-0.5cm}
	\end{center}
\end{figure*}

\begin{figure*}[t!]
	\begin{center}
		\centerline{\includegraphics[width=0.95\linewidth]{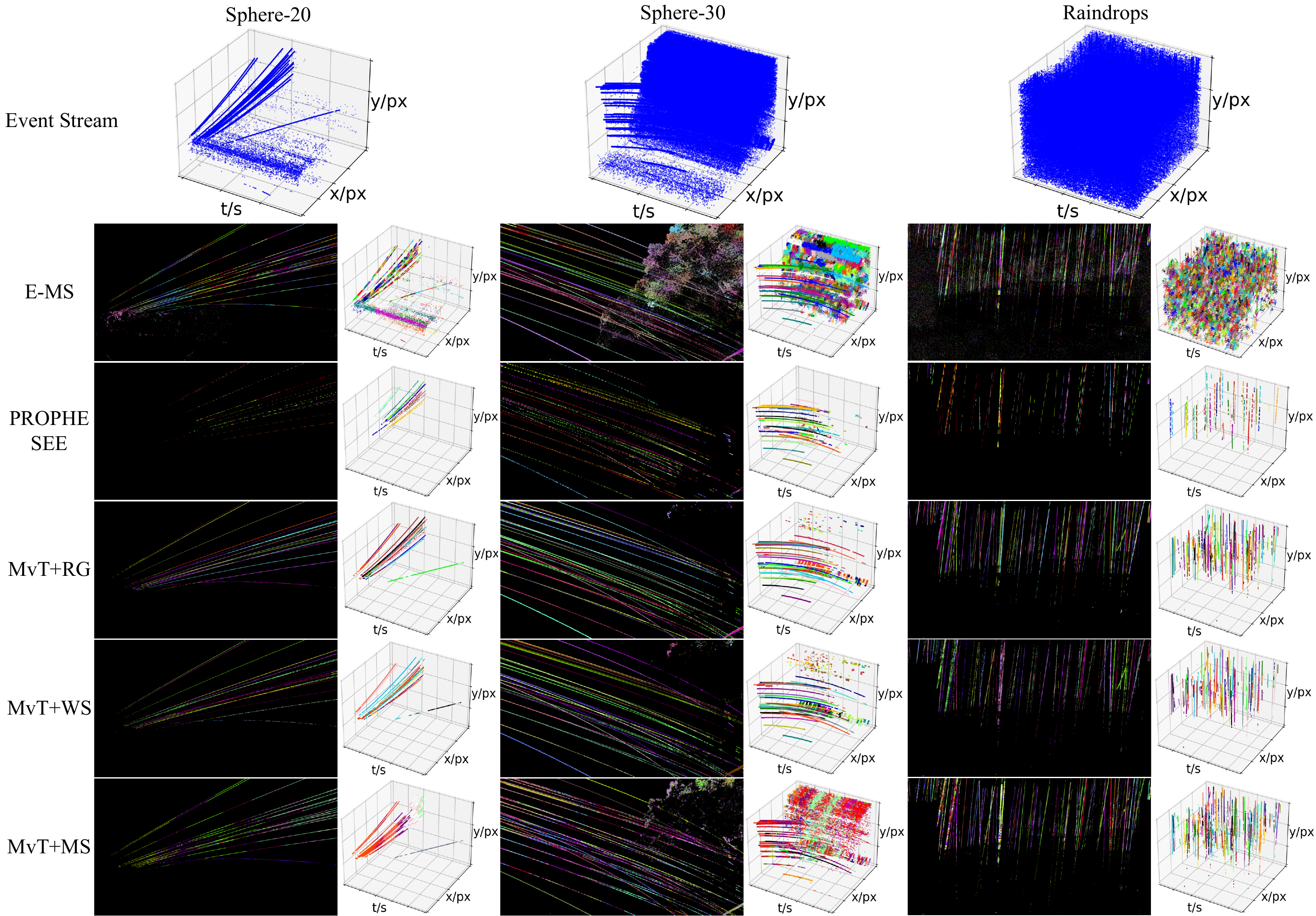}}
		\vspace{-0.2cm}
		\caption{Visualization of tracking results on three scenes in SOD-r. Rows show the 2D and 3D tracking results for three sequences. In the 2D visualizations (XOY views), the background is an event accumulation, where black and white denote the absence and presence of events, respectively. }
		\label{fig12}
		\vspace{-0.7cm}
	\end{center}
\end{figure*}

\textit{(3) Tracking on SOD-r.}
\label{section:SOD}
All methods are evaluated with $B=10$ \textit{ms}. E-MS uses a bandwidth of 0.03, while MvTs are configured with $\Delta \theta=\pi/10$. Fig.~\ref{fig13} presents the FA box plots for the Raindrops sequences, while the remaining results are in the supplementary material. As shown in Fig.~\ref{fig13}, our framework variants rank the top three. In Fig.~\ref{fig12}, MvT+RG and MvT+WS maintain strong performance in both tracking accuracy and stability. However, they also exhibit limitations in trajectory separation and association. When tracklets contain an insufficient number of events, the parameter estimation is unstable or incorrect during the post-processing of views or tracklet association, as observed in Sphere-20 and Sphere-30. In contrast, both MvT+MS and E-MS struggle with these small, densely distributed objects. Despite parameter tuning, their segmentation results tend to incorrectly cluster target events. We attribute this failure to the shared Mean Shift core, which is ill-suited to handle dense distributions. PROPHESEE exhibits moderate performance. It accurately tracks some objects, but resulting trajectories are often too short. These results further validate the effectiveness and generalizability of our framework in tracking small and fast-moving objects compared to methods tailored to specific scenarios. Moreover, the clean outputs of MvTs further demonstrate the framework's ability to mitigate the influence of noise on tracking.

\subsection{Integration with learning-based methods}
\label{sec:lm}
We integrate learning-based methods into MvT to expand its applicability to complex scenarios. Two event-based datasets are used, Ev-UAV and SOD-b. Since object trajectories exhibit distinct spatio-temporal patterns in the YOT and TOX views, appearance-based upstream models perform poorly in these views. Consequently, we adopt an asymmetric strategy, applying learning-based models to the XOY view while utilizing an appearance-independent region growing method for the YOT and TOX. We evaluate six learning-based models, including two image-based models (YOLOX~\cite{geYOLOXExceedingYOLO2021} and SEGMAN~\cite{fuSegMANOmniscaleContext2025}), two event-based models (RVT~\cite{gehrigRecurrentVisionTransformers2023} and EV-SpSegNet~\cite{chenEventbasedTinyObject2025}), and two event-based SOT models (HDETrack~\cite{wangEventStreambasedVisual2024a} and TSETrack~\cite{zhangEfficientVisionTransformer2026}). For MOT comparisons, we select two prevailing image-based trackers, a bounding box-based OCSORT~\cite{caoObservationCentricSORTRethinking2023} and a segmentation-based DEVA~\cite{chengTrackingAnythingDecoupled2023}, besides two event-based SOT models. Since SOD-b is a small-scale set, we only train models on the training set of Ev-UAV and evaluate on SOD-b and the remaining sets of Ev-UAV.

\begin{figure}[t!]
	\begin{center}
		\centerline{\includegraphics[width=1.05\linewidth]{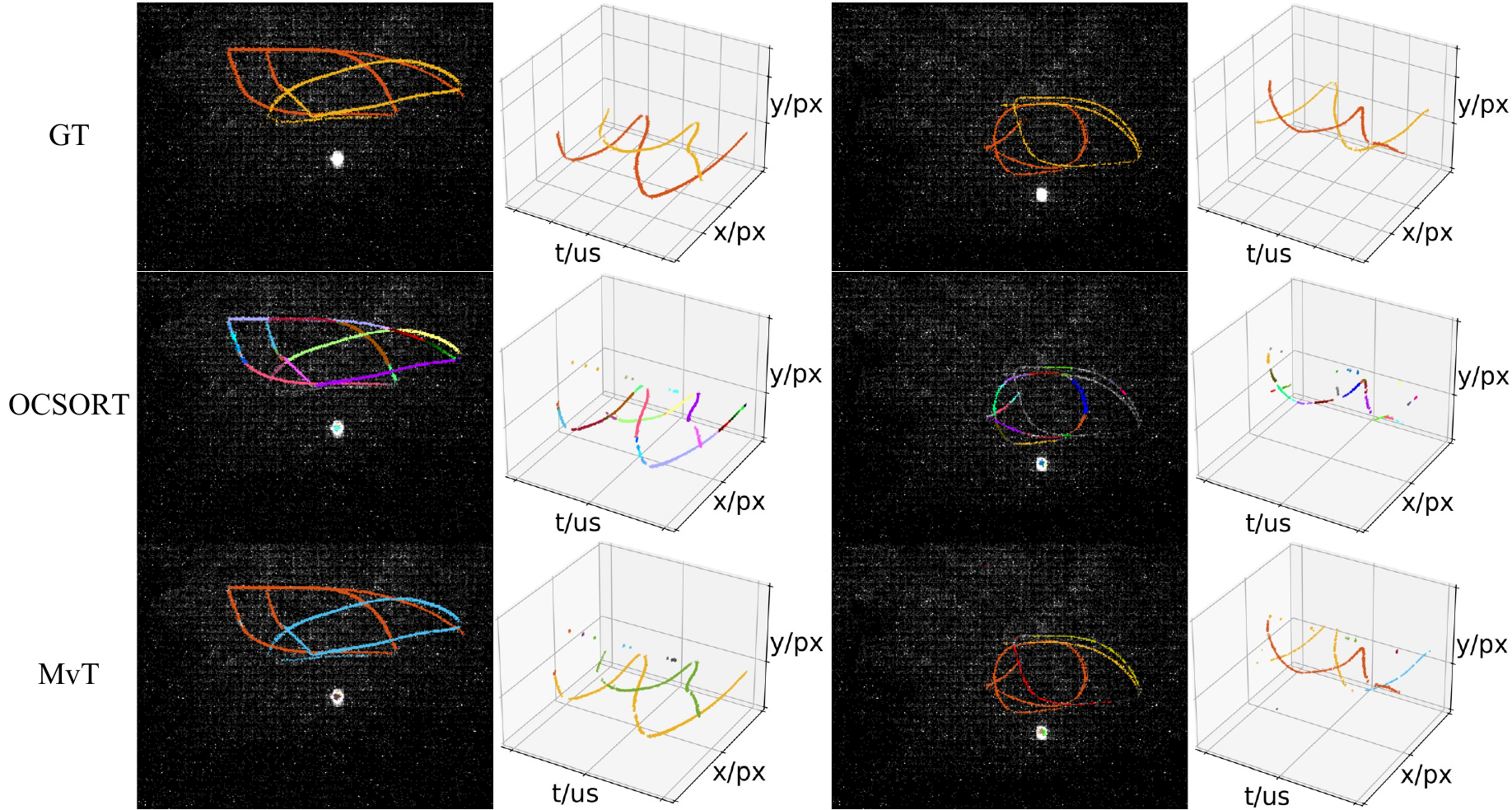}}
		\vspace{-0.2cm}
		\caption{Tracking results of OCSORT and our framework on Ev-UAV based on YOLOX. 2D visualizations are XOY views. The sequences shown in the left and right two columns are test\_017 and val\_019, respectively.}
		\label{fig_evuav_perfm}
		\vspace{-0.6cm}
	\end{center}
\end{figure}

\begin{figure}[t!]
	\begin{center}
		\centerline{\includegraphics[width=1\linewidth]{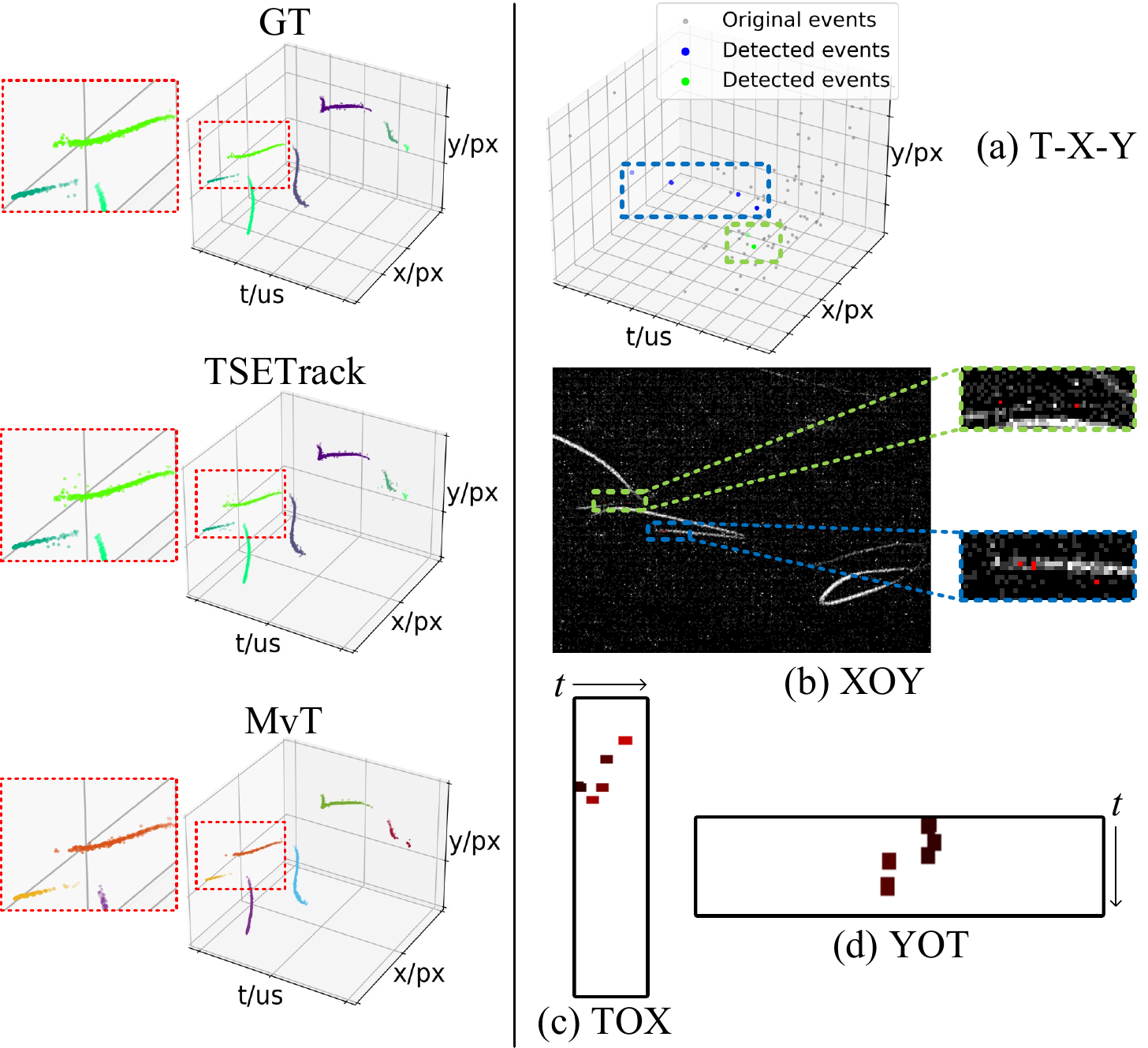}}
		\vspace{-0.2cm}
		\caption{Results on the sequence (test\_013) in Ev-UAV where MvT exhibits a decrease in segmentation performance. The first row presents the GT, TSETrack, and TSETrack+MvT results. In the second row, (a) shows the detected trajectories in the 11th processing window of TSETrack. (b)-(d) present the detections in (a) encoded by our multi-view representation, different colors indicate different tracklets MvT detects.}
		\label{fig_evuav_low_segs}
		\vspace{-0.6cm}
	\end{center}
\end{figure}

\textit{(1) Tracking on Ev-UAV.}
All methods are evaluated with $B=50$ \textit{ms}. Results are reported in Table~\ref{tab8} and Fig.~\ref{fig_evuav_perfm}. Excluding the two SOT models, our method achieves the best and most stable tracking performance across the remaining four upstream models. Specifically, integrating MvT with YOLOX and RVT outperforms the detection-based OCSORT by over 15\% Track IoU, 7\% Track Acc, and 10\% IDF1, and surpasses the segmentation-based DEVA by over 14\% Track IoU, 28\% Track Acc, and 43\% IDF1. However, our method yields lower segmentation performance than the original upstream models. This degradation occurs because the multi-view strategy recognizes scattered events around trajectories as outliers. As illustrated in the first row of Fig.~\ref{fig_evuav_low_segs}, TSETrack+MvT classifies surrounding events as outliers due to their irregular distribution in the TOX view (Fig.~\ref{fig_evuav_low_segs}(c)). Although this discards some valid trajectory points, the resulting trajectories remain smooth and trackable. The two SOT models, HDETrack and TSETrack, achieve IDF1 scores of nearly 100\%. Their low false positives and false negatives can be attributed to supervised training, GT initialization, and the small number of tracking targets. To evaluate the generalization of these methods, we conduct further experiments on SOD-b without fine-tuning.

Regarding computational overhead, MvT operates at 146-215 FPS (Table~\ref{tab8}) when paired with bounding-box-based detectors, HDETrack, TSETrack, YOLOX and RVT. For segmentation networks, MvT still achieves 19-22 FPS, which is 6$\times$ faster than DEVA under SEGMAN. These results show the efficiency of our framework.

\begin{table*}[t]
	\begin{center}
		\renewcommand\arraystretch{1}
		\setlength{\abovecaptionskip}{0.1cm}
		\vspace{-0.5cm}
		\caption{Results on SOD-b. Hereafter, only the unique best values are highlighted in bold.} 
		\label{tab9}
		\begin{tabular}{
			>{\centering}p{1.6cm}|
			>{\centering}p{1.5cm}|
			>{\centering}p{1.5cm}|
			>{\centering}p{1.5cm}|
			>{\centering}p{1.6cm}|
			>{\centering}p{1.6cm}|
			>{\centering}p{1.7cm}|
			>{\centering\arraybackslash}p{1.2cm}
			}
			\Xhline{1.2pt}
			Method   & Task Type    & Event Rep.  & Seg. IoU(\%)  & Seg. Acc(\%)   & Track IoU(\%) & Track Acc.(\%) & IDF1(\%)       \\
			\Xhline{1.2pt}
			HDETrack & SOT          & Event Count & {0.70}        & {24.90}        & {0.60}        & {0.73}         & {13.33}        \\
			+MvT     & MOT          & MvP         & {0.70}        & {24.90}        & \textbf{1.69} & \textbf{9.68}  & \textbf{38.38} \\
			\hline
			TSETrack & SOT          & Event Count & 0.82          & \textbf{29.14} & 1.23          & 1.55           & 13.64          \\
			+MvT     & MOT          & MvP         & {0.82}        & {29.13}        & \textbf{2.78} & \textbf{9.83}  & \textbf{33.16} \\
			\hline
			YOLOX*   & Detection    & Event Count & 0.41          & 0.42           & --            & --             & --             \\
			+OCSORT* & MOT          & MvP         & \textbf{0.42} & 0.42           & \textbf{2.87} & 0.40           & 6.52           \\
			+MvT     & MOT          & MvP         & 0.41          & {0.42}         & 2.64          & \textbf{0.41}  & \textbf{8.51}  \\
			\hline
			SEGMAN*  & Segmentation & Event Count & {0.08}        & {0.12}         & --            & --             & --             \\
			+DEVA*   & MOT          & MvP         & 0.00          & 0.00           & 0.00          & 0.00           & 0.00           \\
			+MvT     & MOT          & MvP         & {0.08}        & {0.12}         & \textbf{0.35} & \textbf{0.09}  & \textbf{5.29}  \\
			\Xhline{1.2pt}
		\end{tabular}
		\vspace{-0.6cm}
	\end{center}
\end{table*}

\textit{(2) Tracking on SOD-b.}
All methods are evaluated with $B=16.67$ \textit{ms}. Without training, several models fail to produce valid results, so we integrate the remaining four upstream models into MvT for evaluation. In Table~\ref{tab9}, all methods undergo a substantial performance drop due to severe domain shift. Since our method is a tracking framework that relies on detections, its absolute tracking performance is inherently bounded by the degraded detection accuracy. Even under such low accuracy, MvT still achieves the best relative tracking performance, surpassing HDETrack and TSETrack by over 7\% in Track Acc and 20\% in IDF1. These improvements further confirm the generalization and effectiveness of our multi-view framework in complex motion scenarios.

\subsection{Case Study}
\label{sec:case}
\begin{figure}[t!]
	\begin{center}
		\vspace{-0.5cm}
		\centerline{\includegraphics[width=1\linewidth]{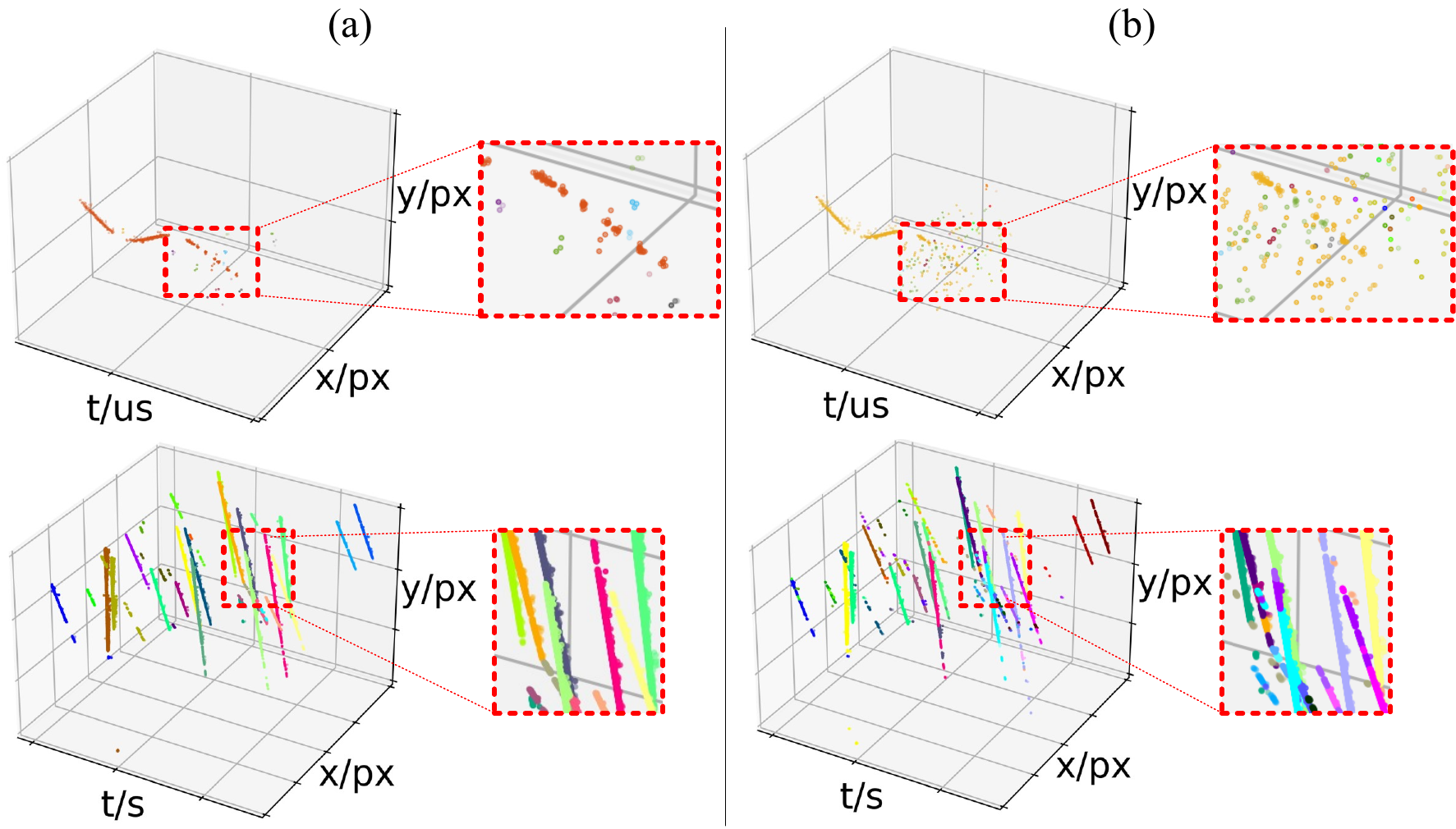}}
		\vspace{-0.2cm}
		\caption{Case study comparing existing pseudo-frames and multi-view representation under occlusion (first row, Ev-UAV test\_007 sequence) and closely distributed trajectories (second row, SOD-r Raindrops). (a) shows the results of multi-view variants, MvT+TSETrack and MvT+RG. (b) shows the results of voxel-grid variant MvF+TSETrack and single-view variant MvF+RG.}
		\label{fig_case_study}
		\vspace{-0.6cm}
	\end{center}
\end{figure}

This study is to validate the tracking disambiguation of our framework compared to the pseudo-frame representation. In Fig.~\ref{fig_case_study}, we present two cases involving occluded and closely distributed trajectories. Sub-figure (a) shows the results of MvT+TSETrack/RG and sub-figure (b) shows the results of MvF+TSETrack/RG. Since TSETrack is a voxel-grid-based model and RG operates on a single view, we integrate our MvF module with them to construct the voxel-grid variant and the single-view variant, denoted as MvF+TSETrack and MvF+RG, respectively. In the first row of Fig.~\ref{fig_case_study}, the single-view variant fails to distinguish trajectories from occlusions, whereas our multi-view model succeeds. For closely distributed objects, the single-view method cannot determine whether the events mapped to a pixel originate from one object. This causes several trajectory segments with the same color to be discretely distributed in the T-X-Y space. In contrast, our multi-view representation effectively separates them.

\subsection{Gaussian Pyramid Design}
\label{sec:gauss}

\begin{figure}[t!]
	\begin{center}
		\vspace{-0.5cm}
		\centerline{\includegraphics[width=1\linewidth]{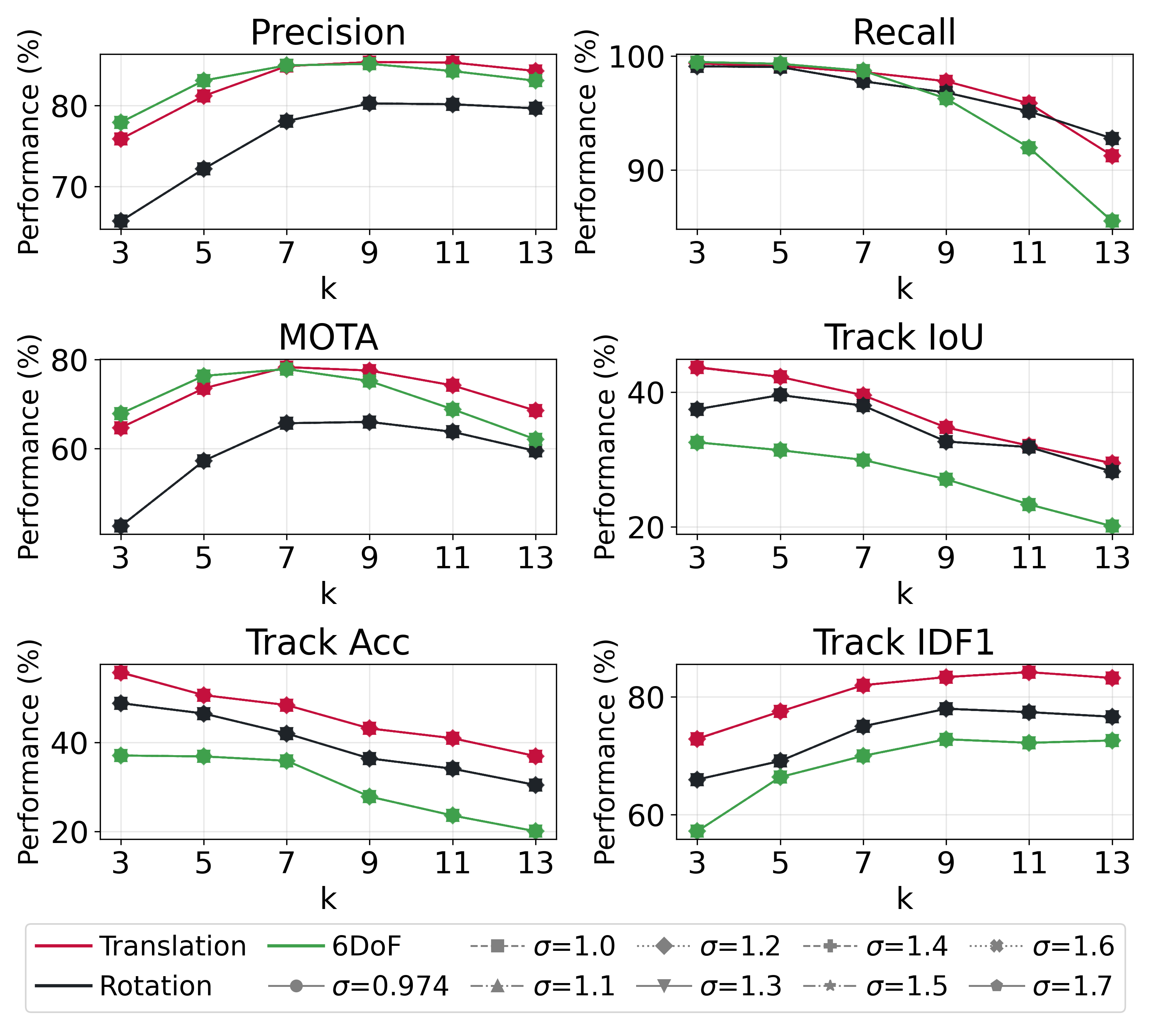}}
		\vspace{-0.3cm}
		\caption{The results of different combinations of $\sigma$ and $k$ in Gaussian kernel design using RG on ECD.}
		\label{Sig_K_ablation_ECD_RG}
		\vspace{-0.6cm}
	\end{center}
\end{figure}

\begin{figure}[t!]
	\begin{center}
		\centerline{\includegraphics[width=1.\linewidth]{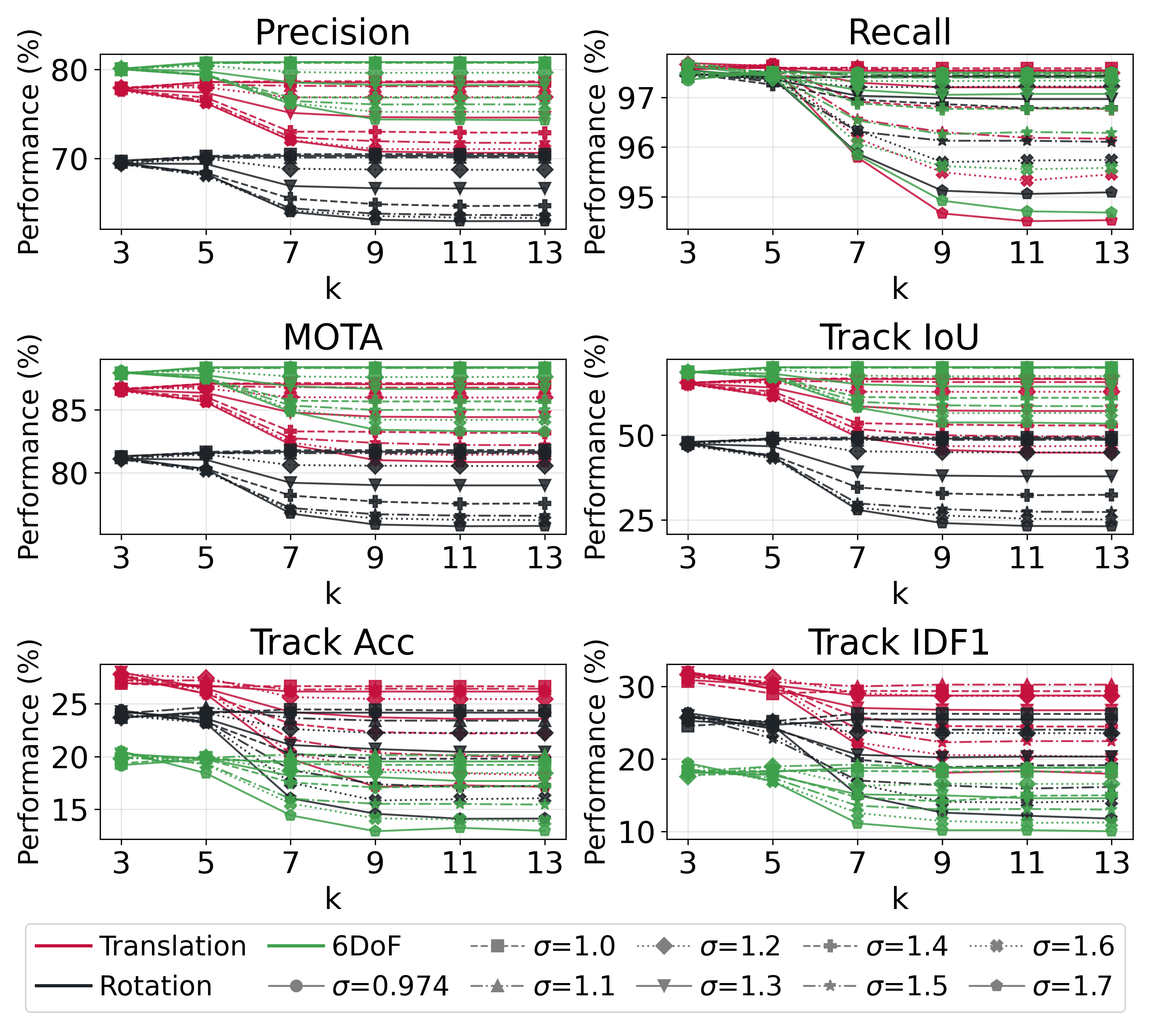}}
		\vspace{-0.3cm}
		\caption{The results of different combinations of $\sigma$ and $k$ in Gaussian kernel design using WS on ECD.}
		\label{Sig_K_ablation_ECD_WS}
		\vspace{-0.6cm}
	\end{center}
\end{figure}

The Gaussian pyramid is a crucial component in our MvP module, designed to mitigate the pattern discretization of event distributions in views. The design of Gaussian kernels and the number of layers are essential for its effectiveness. We test a series of values for these parameters to determine the best combination. The kernel design involves two parameters, kernel size $k$ and standard deviation $\sigma$ in Gaussian kernels. We initially set the downsampling scale $s=0.8$, and the stopband edge gain $\alpha=26$dB. Then the lower bound of $\sigma$ for anti-aliasing is $\sigma_{anti} = 0.974$. The tested values of $\sigma$ and $k$ are $\sigma \in \{0.974, 1.0, 1.1, 1.2, 1.3, 1.4, 1.5, 1.6, 1.7\}$, and $k \in \{3, 5, 7, 9, 11, 13\}$. We evaluate parameter sensitivity on the ECD dataset using RG (Fig.~\ref{Sig_K_ablation_ECD_RG}) and WS (Fig.~\ref{Sig_K_ablation_ECD_WS}). Fig.~\ref{Sig_K_ablation_ECD_RG} reveals that variations in $\sigma$ have almost no impact on final tracking performance. This is because RG identifies components based on non-zero pixel connectivity ($>0$ for the target and $=0$ for the background). Under this binary criterion, the Gaussian kernel with continuous weighting acts like a box filter. However, as $k$ spans $[3, 13]$, performance fluctuations become noticeable with a maximum of 15\%.

\begin{figure}[t!]
	\begin{center}
		\centerline{\includegraphics[width=1.02\linewidth]{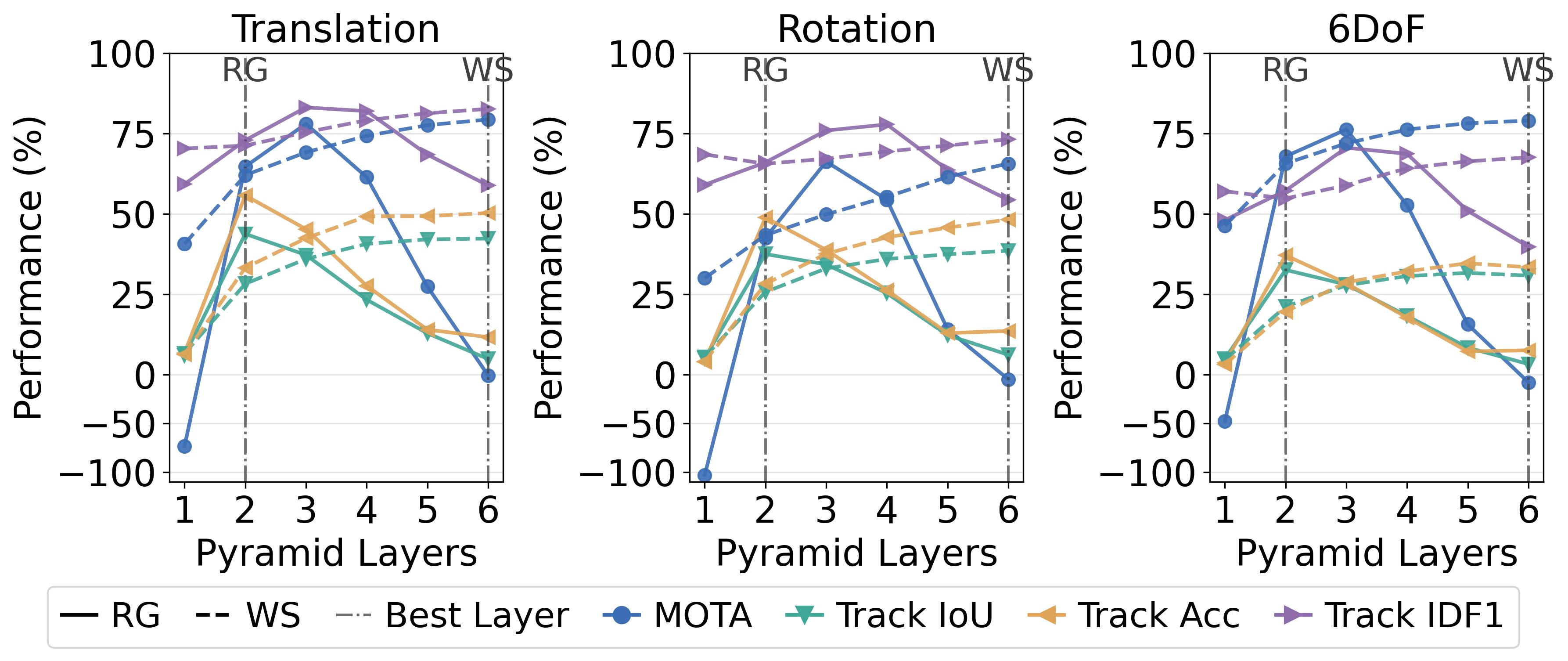}}
		\vspace{-0.3cm}
		\caption{Results of different numbers of layers on Gaussian pyramid.}
		\label{figure_GaussPyLvl}
		\vspace{-1cm}
	\end{center}
\end{figure}

In Fig.~\ref{Sig_K_ablation_ECD_WS}, when $k=3$, all $\sigma$ values perform similarly well, with slightly higher Recall, Track Acc, and Track IDF1 when $\sigma \ge 1.4$. Unlike RG, when $k\ge5$, WS achieves its best performance when $\sigma$ lies in $[0.974, 1.1]$, which is around the anti-aliasing lower bound $\sigma_{anti}$. As $\sigma$ grows, performance degrades significantly. This discrepancy stems from their distinct mechanisms, as WS relies more heavily on structural patterns. To comprehensively accommodate different image-based techniques, the Gaussian kernel is configured as $k=3$ and $\sigma=1.6$. Although $k=3$ does not yield the optimal trade-off for RG, it clearly achieves the best performance for WS. Larger kernels may suppress more noise and connect broken tracklets, yielding higher Precision, but they cause noticeable performance fluctuations and risk over-smoothing adjacent trajectory patterns. This indicates that designing a dynamic kernel size selection strategy is a meaningful direction for future work, as discussed in Sec.~\ref{section:discussion}. The Gaussian filtering is implemented using OpenCV, with the kernel normalized to preserve the DC component.

With respect to the number of layers, we evaluate configurations from 1 to 6, where one layer indicates that no pyramid is applied. As shown in Fig.~\ref{figure_GaussPyLvl}, using only one layer significantly degrades the performance of both RG and WS, confirming the necessity of the Gaussian pyramid. RG achieves the optimal result with 2 layers, whereas WS performs best with 6 layers. This difference also stems from their distinct mechanisms. Furthermore, the WS results verify that the scale space built by the Gaussian pyramid helps mitigate pattern discretization. Because RG is more sensitive to the number of layers than WS, a 2-layer configuration is selected as the default.

\begin{table}[H]
	\begin{center}
		\renewcommand\arraystretch{1}
		\vspace{-0.3cm}
		\setlength{\abovecaptionskip}{0.1cm}
		\caption{Ablation results of the multi-view representation on ECD using MS. }
		\label{tab3}
		\begin{footnotesize}
			\begin{tabular}{
				>{\centering}p{1.cm}|
				>{\centering}p{1cm}|
				>{\centering}p{1.cm}|
				>{\centering}p{1.cm}|
				>{\centering}p{1.1cm}|
				>{\centering\arraybackslash}p{1.cm}
				}
				\Xhline{1.2pt}
				{Seq.}                     & Rep.          & P(\%)          & R(\%)          & MOTA(\%)       & IDF1(\%)       \\

				\Xhline{1.2pt}
				\multirow{3}*{translation} & MvP           & \textbf{76.49} & \textbf{98.91} & \textbf{64.49} & \textbf{75.87} \\
				~                          & SingV         & 74.22          & 98.83          & 61.42          & 70.65          \\
				~                          & VG-\ding{172} & 30.21          & 97.98          & -173.50        & 12.15          \\
				~                          & VG-\ding{173} & 10.62          & 64.96          & -529.21        & 2.05           \\
				\hline
				\multirow{3}*{rotation}    & MvP           & \textbf{69.89} & {98.45}        & \textbf{49.56} & \textbf{70.98} \\
				~                          & SingV         & 65.42          & \textbf{98.61} & 41.14          & 65.16          \\
				~                          & VG-\ding{172} & 27.86          & 97.73          & -200.76        & 14.09          \\
				~                          & VG-\ding{173} & 10.09          & 77.01          & -669.07        & 1.87           \\
				\hline
				\multirow{3}*{6dof}        & MvP           & \textbf{77.18} & {98.80}        & \textbf{64.73} & \textbf{57.94} \\
				                           & SingV         & 76.90          & \textbf{98.95} & 63.45          & 57.13          \\
				~                          & VG-\ding{172} & 28.78          & 98.41          & -193.71        & 6.03           \\
				~                          & VG-\ding{173} & 10.58          & 51.60          & -417.65        & 1.41           \\
				\Xhline{1.2pt}
			\end{tabular}
		\end{footnotesize}
		\vspace{-0.4cm}
	\end{center}
\end{table}

\begin{table}[H]
	\begin{center}
		\renewcommand\arraystretch{1}
		\vspace{-0.3cm}
		\setlength{\abovecaptionskip}{0.1cm}
		\caption{Ablation results of the multi-view representation on Ev-UAV.}
		\label{tab3_5}
		\begin{tabular}{
			>{\centering}p{1.3cm}|
			>{\centering}p{1cm}|
			>{\centering}p{1.6cm}|
			>{\centering}p{1.65cm}|
			>{\centering\arraybackslash}p{1cm}
			}
			\Xhline{1.2pt}
			Method               & Rep.          & Track IoU(\%)  & Track Acc(\%)  & IDF1(\%)       \\

			\Xhline{1.2pt}
			\multirow{2}*{YOLOX} & MvP           & \textbf{50.79} & \textbf{54.44} & \textbf{71.74} \\
			                     & SingV         & 34.20          & 45.27          & 61.73          \\
			\hline
			\multirow{2}*{RVT}   & MvP           & \textbf{33.04} & \textbf{13.99} & 39.22          \\
			                     & VG-\ding{174} & 18.31          & 11.67          & \textbf{44.37} \\
			\Xhline{1.2pt}
		\end{tabular}
		\vspace{-0.4cm}
	\end{center}
\end{table}

\subsection{Ablation Study}
\label{sec:AS}
We conduct ablation studies on two components, multi-view representation and time realignment.

\textit{(1) Multi-view representation}. We evaluate our multi-view representation (MvP) against a voxel-grid-based representation (VG, $\in \mathbb{R}^{H\times W\times T}$, $T = 200$) and a single-view representation (SingV, $\in \mathbb{R}^{H\times W}$). While MvP and SingV share a 2D frame format, VG is inherently 3D data. To comprehensively evaluate VG against MvP, we design four experiments addressing different VG processing paradigms. First, project the 3D VG onto the same orthogonal views (XOY, YOT, TOX) used in MvP. This process is mathematically equivalent to applying MvP without the Time Realignment (TR) module. As shown in the subsequent ablation study (Table~\ref{tab4} and Table~\ref{tab5}), MvP consistently outperforms this VG-projection variant. Second, apply the 3D Mean Shift to VG. Third, treat VG as a multi-channel 2D image, where the same 2D Mean Shift is applied independently to each channel. Fourth, utilize an event-based VG detector, RVT, as the upstream model to compare tracking performance under the VG and MvP pipelines. The latter three experiments, denoted as VG-\ding{172}, VG-\ding{173}, and VG-\ding{174}, are conducted on ECD and Ev-UAV, with results in Table~\ref{tab3} and Table~\ref{tab3_5}. All variants are evaluated under identical parameter settings.

In Table~\ref{tab3}, the MvP variant significantly outperforms two VG variants across all metrics and leads the SingV variant in most metrics. In Table~\ref{tab3_5}, the MvP variant again outperforms the VG variant by a large margin, achieving improvements of over 14\% in Track IoU and 2\% in Track Acc. This confirms that MvP successfully isolates true trajectories from adjacent trajectories or background noise. We also observe a drop in IDF1 for the MvP variant. This is because in the VG model, spatially close events are often merged into one trajectory, artificially maintaining ID consistency. In contrast, the MvP variant distinguishes individual tracklets across multiple views, inevitably increasing the number of identities and lowering IDF1, as illustrated in Fig.~\ref{fig_case_study}. Furthermore, we integrate an image-based detector, YOLOX, to evaluate the MvP and SingV representations on Ev-UAV. The MvP model outperforms the SingV model by 16\% in Track IoU, 9\% in Track Acc, and 10\% in IDF1, further confirming the effectiveness of our representation.

\textit{(2) Time realignment}. Time realignment aims to mitigate the effects of temporal discretization in subsequent frame-based processing. We evaluate MvT+RG with and without this module, as reported in Table~\ref{tab4} and Table~\ref{tab5}. The model with time realignment consistently surpasses its counterpart across all metrics, indicating the effectiveness of this module in processing sparsely and discretely distributed events.

\begin{table}[H]
	\begin{center}
		\renewcommand\arraystretch{1}
		\vspace{-0.3cm}
		\setlength{\abovecaptionskip}{0.1cm}
		\caption{Ablation results of time realignment (TR) on ECD.}
		\label{tab4}
		\begin{tabular}{
			>{\centering}p{1.cm}|
			>{\centering}p{1cm}|
			>{\centering}p{1.cm}|
			>{\centering}p{1.cm}|
			>{\centering}p{1.1cm}|
			>{\centering\arraybackslash}p{1.cm}
			}
			\Xhline{1.2pt}
			{Seq.}                     & {Model} & P(\%)          & R(\%)          & MOTA(\%)       & IDF1(\%)       \\

			\Xhline{1.2pt}
			\multirow{2}*{translation} & MvT+RG  & \textbf{75.85} & \textbf{99.36} & \textbf{64.67} & \textbf{72.84} \\
			~                          & w/o TR  & 75.80          & 99.33          & 64.47          & 72.23          \\
			\hline
			\multirow{2}*{rotation}    & MvT+RG  & 65.73          & {99.11}        & {42.47}        & \textbf{65.97} \\
			~                          & w/o TR  & 65.73          & 99.11          & 42.47          & 65.81          \\
			\hline
			\multirow{2}*{6dof}        & MvT+RG  & {77.90}        & {99.48}        & \textbf{67.83} & \textbf{57.21} \\
			~                          & w/o TR  & 77.90          & 99.48          & 67.81          & 56.52          \\
			\Xhline{1.2pt}
		\end{tabular}
		\vspace{-0.4cm}
	\end{center}
\end{table}

\begin{table}[H]
	\begin{center}
		\renewcommand\arraystretch{1}
		\vspace{-0.3cm}
		\setlength{\abovecaptionskip}{0.1cm}
		\caption{Ablation results of time realignment (TR) on Ev-UAV.}
		\label{tab5}
		\begin{tabular}{
			>{\centering}p{1.3cm}|
			>{\centering}p{1cm}|
			>{\centering}p{1.6cm}|
			>{\centering}p{1.65cm}|
			>{\centering\arraybackslash}p{1cm}
			}
			\Xhline{1.2pt}
			Method                  & Model  & Track IoU(\%)  & Track Acc(\%)  & IDF1(\%)       \\
			\Xhline{1.2pt}
			\multirow{2}*{HDETrack} & MvT+RG & \textbf{60.91} & \textbf{67.17} & \textbf{73.15} \\
			~                       & w/o TR & 59.19          & 66.64          & 71.66          \\
			\hline
			\multirow{2}*{TSETrack} & MvT+RG & \textbf{60.17} & \textbf{67.49} & \textbf{75.14} \\
			~                       & w/o TR & 59.14          & 67.45          & 74.52          \\
			\hline
			\multirow{2}*{RVT}      & MvT+RG & \textbf{33.04} & \textbf{13.99} & \textbf{39.22} \\
			~                       & w/o TR & 33.02          & 13.73          & 38.82          \\
			\Xhline{1.2pt}
		\end{tabular}
		\vspace{-0.4cm}
	\end{center}
\end{table}

\section{Discussion}
\label{section:discussion}
Comprehensive evaluations across four datasets featuring objects of varying scales and motion patterns demonstrate the effectiveness, generalization, and flexibility of the MvT framework. By integrating with diverse upstream detection or segmentation models, MvT successfully adapts to a wide range of tracking environments.

However, as the association model of our method relies on motion continuity rather than appearance features or specifically learned motion models, it may struggle during prolonged occlusions accompanied by significant directional changes. Specifically, while $\Delta \theta$ in association \textbf{Criterion~\ref{Cr2}} bounds directional calculation errors, it restricts the motion model to near-linear assumptions. The optimal $\Delta \theta$ may vary for objects with different motion patterns. Future work will explore adaptive methods to dynamically determine $\Delta \theta$ based on object characteristics or complex motion models. Currently, MvT utilizes only the spatio-temporal coordinates ($t$, $x$, $y$) of each event. As introduced in Sec.~\ref{section:Introd}, some researchers~\cite{paredes-vallesBackEventBasics2021, scheerlinckFastImageReconstruction2020} have leveraged event polarity to reconstruct intensity images containing rich texture and appearance information. Integrating polarity and its underlying intensity gradient field to infer object motion represents a promising direction for adaptively determining $\Delta \theta$. Lastly, the isotropic Gaussian kernel used in MvP may not be optimal for all views, as pattern discontinuities vary across projection views. Furthermore, our Gaussian kernel experiments show that optimal kernel designs also differ across objects. It is a worthwhile direction for future to explore adaptive kernel shapes and sizes to better mitigate pattern discontinuities, especially in cases of complex motion combinations or varying object sizes.

\section{Conclusion}
\label{section:conclusion}
In this work, we have presented MvT, a general framework that leverages an intuitive multi-view representation to effectively encode spatio-temporal information from event streams for event-based tracking.
Our representation enables the direct application of image-based upstream detection or segmentation algorithms while leveraging complementary dimensional information to resolve tracking ambiguities, such as occlusion and overlap, in a single view. An association model consisting of three criteria under different scenarios further ensures robust tracklet linking and stable performance. Comprehensive experiments across four datasets demonstrate that MvT can flexibly integrate with learning-based and non-learning-based models to reliably track objects of varying sizes, outperforming existing event-based methods. Crucially, as a training-free approach, MvT enables robust tracking even in extreme scenarios where GT annotations are unobtainable, positioning it as a practical and generalizable framework for event-based tracking in real-world applications.

\bibliographystyle{IEEEtran}
\balance
\bibliography{refs}

\vfill
\end{document}